\documentclass[10pt,twocolumn,letterpaper]{article}

\usepackage{paper}

\usepackage{amsmath,amssymb}
\usepackage{booktabs}
\usepackage{multirow}
\usepackage{subcaption}
\usepackage{xspace}
\usepackage{graphicx}
\usepackage{enumitem}
\usepackage{cuted}        

\newcommand{\method}{GATS-WM\xspace}
\newcommand{\gats}{GATS\xspace}

\DeclareMathOperator*{\argmin}{arg\,min}

\definecolor{linkblue}{rgb}{0.21,0.49,0.74}
\usepackage[pagebackref,breaklinks,colorlinks,allcolors=linkblue]{hyperref}

\title{Let LiDAR Decide Where to Look: Geometry-Aware\\Token Selection for Efficient Sparse Latent World Models}

\author{
\begin{tabular}{c@{\hspace{4em}}c}
Zeping Zhang & Zhuoya Zhao \\
University of Ottawa & University of Ottawa \\
{\tt\small Zeping.Zhang@uottawa.ca} & {\tt\small zoezhao0427@gmail.com}
\end{tabular}
\\[1em]
Robert Lagani\`ere \\
University of Ottawa \\
{\tt\small laganier@uottawa.ca}
}

\begin{document}
\maketitle
\begin{strip}
\centering
\includegraphics[width=\linewidth]{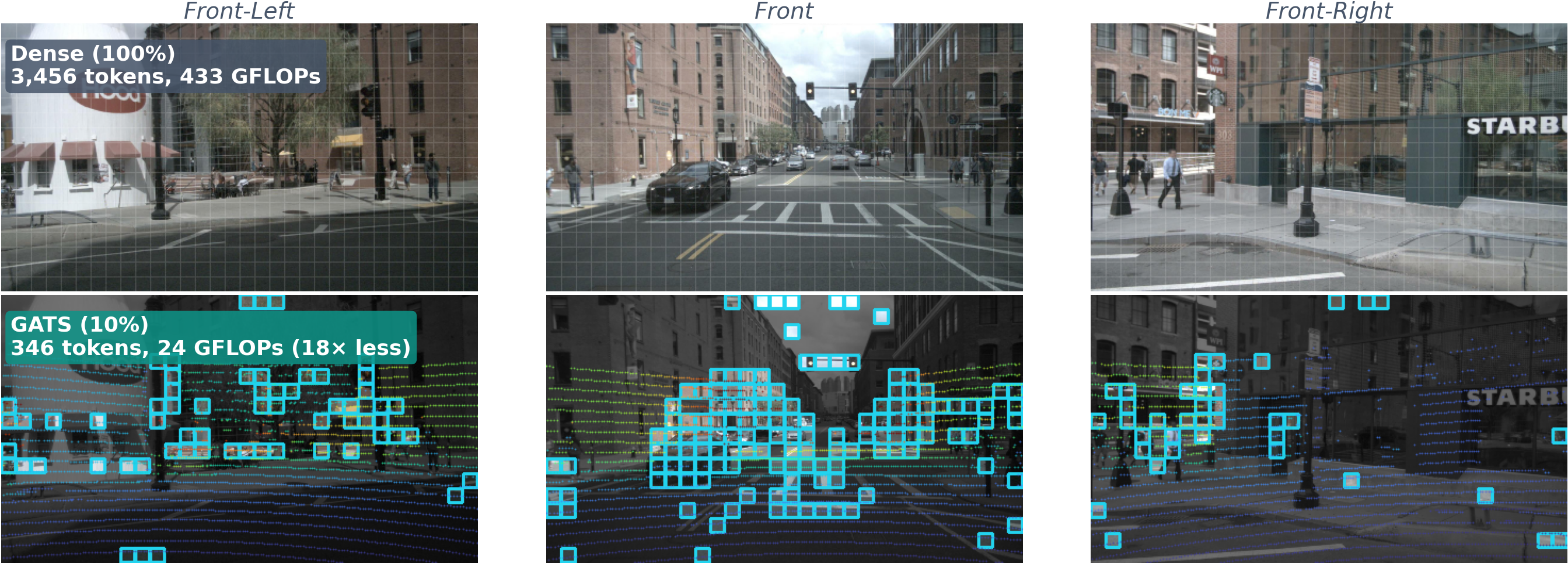}
\captionsetup{hypcap=false}
\captionof{figure}{\textbf{Dense vs.\ sparse token processing.} \emph{Top:} A standard pipeline processes all 3{,}456 image tokens across 6 cameras (433 GFLOPs end-to-end). \emph{Bottom:} \method projects LiDAR points onto the image grid, scores each patch, and outputs the indices of the top-$K$ to retain.
We show the 10\% budget for visual clarity.
At 10\%, only 346 tokens are encoded, reducing compute to 24 GFLOPs ($18\times$ less). Selected regions cover driving-critical content—vehicles, pedestrians, and the drivable road surface—while sky and irrelevant background elements are pruned.}
\captionsetup{hypcap=true}
\label{fig:teaser}
\end{strip}

\begin{abstract}
Vision-centric autonomous driving models encode thousands of image tokens per frame, yet most depict sky, buildings, or empty road---content irrelevant to planning.
Worse, these tokens dilute the planner's attention: we show that randomly discarding 75\% of tokens already outperforms the dense model, because the planner can concentrate on the remaining driving-critical content.
Building on this finding, we present \textbf{Geometry-Aware Token Selection (\gats)}, which uses LiDAR and radar point projections to select a small fraction of image tokens \emph{before} the vision backbone runs.
Because the selector operates on measured 3D geometry rather than visual features, the backbone encodes only the selected patches, cutting end-to-end FLOPs by up to $18\times$.
Each kept token receives a 3D anchor from its LiDAR point and is fed to a self-supervised latent world model that predicts future features and 3D displacements without box annotations.
At 25\% budget, \gats achieves a 3-sec L2 displacement error of 0.520\,m on nuScenes and an EPDMS of 88.9 on NAVSIM---surpassing the dense baseline (0.573\,m / 85.9) at $6.6\times$ fewer total FLOPs.
At an extreme 10\% budget (24 GFLOPs, $18\times$ reduction), it still beats the dense baseline on both nuScenes (0.537\,m vs.\ 0.573\,m L2) and NAVSIM (87.9 vs.\ 85.9 EPDMS).
On 739 intent-critical scenarios, \gats covers 95.8\% of hazardous vehicles and achieves a 0.09\% collision rate---73.5\% lower than the dense model (0.34\%).

\end{abstract}

\section{Introduction}
\label{sec:intro}

Autonomous driving demands real-time 3D understanding, motion forecasting, and trajectory planning.
Latent world models~\cite{lecun2022path,bardes2024vjepa,garrido2024dinowm} learn to predict future scene states in a compact feature space and can train against frozen self-supervised targets such as DINOv2~\cite{oquab2024dinov2}, capturing dynamics without manual annotations.

These models, however, treat every image patch equally.
Six surround cameras encoded by DINOv2-Small produce ${\sim}$3{,}500 tokens per frame at 185 GFLOPs, and the downstream planner must attend to all of them.
Most tokens depict sky, irrelevant background elements, or road far from the ego path.
They carry little planning-relevant information yet dominate both the compute budget and the planner's finite attention capacity.
This raises a natural question: can we determine which tokens matter before encoding them?

\paragraph{The attention-dilution bottleneck.}
Adding more tokens or richer features does not help.
In fact, randomly discarding 75\% of tokens already produces better planning than the dense model (a 3-sec displacement error of 0.550\,m vs.\ 0.573\,m L2, \cref{tab:main_nusc}).
The dense model feeds 6{,}913 tokens to the planner's cross-attention, yet a nearby vehicle occupies only 3 tokens (median across all val-set vehicles); the planning signal is diluted by thousands of irrelevant patches.
Multimodal fusion~\cite{liu2023bevfusion,chitta2023transfuser,li2025wote} enriches the quality of scene representations, but does not address this dilution---nor does it reduce the cost of the dense visual encoding pass.
A controlled ablation confirms this: adding a learned LiDAR-feature fusion branch (4-layer MLP, 2.4M parameters) to our sparse model does not improve planning L2 (\cref{sec:ablation}).
The bottleneck is not what the model sees but how much irrelevant content competes for the planner's attention.

\paragraph{Key insight: geometry-guided visual compute routing.}
L4-class autonomous vehicles typically carry LiDAR and radar alongside cameras, which directly measure where 3D structure exists.
A LiDAR return above ground level signals a relevant obstacle; multi-sweep motion residuals flag moving agents; radar provides radial velocity.
These cues are computable from a simple point-cloud projection ($O(N_\text{pts})$) before any image feature is extracted.
The resulting token scores are fed through lightweight MLPs trained end-to-end with the world-model loss.

Two consequences follow.
\textbf{(1)}~Because selection depends only on 3D measurements, the ViT backbone can encode only the selected patches---something impossible when selection relies on backbone outputs.
\textbf{(2)}~Measured geometry is not biased by training-set frequency: any vehicle that reflects LiDAR above the ground plane is selected, regardless of how often it appears during training.
This matters for rare but safety-critical scenarios where learned selectors fail (\cref{sec:intent}).

\paragraph{Contributions.}
We present \textbf{\method} (Geometry-Aware Token Selection for World Models), a sparse latent world model with four contributions:

\begin{enumerate}[leftmargin=*,topsep=2pt,itemsep=1pt]
\item \textbf{Geometry-aware token selection (\gats).} A three-stream scorer uses LiDAR geometry, motion residuals, and radar velocity to select a fixed fraction (e.g. 10\% or 25\%) of image tokens.
The selector operates on 3D measurements only and does not require image features at inference.

\item \textbf{3D-anchored self-supervised world model.} Each selected token is anchored at the median 3D position of its LiDAR points, enabling cross-camera association, ego-motion-compensated temporal alignment (history anchors are transformed to the current ego frame via known odometry), and self-supervised motion prediction via feature-geometry matching---all without box annotations.

\item \textbf{Safety-critical coverage.} On 739 mined intent-critical scenarios, \gats covers 95.8\% of hazardous vehicles---far above random (81.5\%) and learned selectors (70.0\%)---while achieving a collision rate of 0.09\%, 73.5\% lower than the dense model.

\item \textbf{End-to-end efficiency.} At 25\% budget, \gats achieves 0.520\,m L2 (nuScenes) and 88.9 (NAVSIM)---both better than the dense baseline, but at $6.6\times$ fewer total FLOPs.
At 10\% budget, total cost drops to 24 GFLOPs ($18\times$ reduction) while still surpassing the dense model on both nuScenes and NAVSIM (87.9 vs.\ 85.9).

\end{enumerate}

\section{Related Work}
\label{sec:related}

\paragraph{Latent world models for driving.}
Pixel-space world models~\cite{hu2023gaia,wang2024drivedreamer,gao2024vista} generate future images but are expensive and entangle appearance with dynamics.
Latent alternatives predict in a learned feature space: MILE~\cite{hu2022mile} compresses BEV features into a compact probabilistic latent; MUVO~\cite{bogdoll2024muvo} predicts multimodal sensor futures; Think2Drive~\cite{li2024think2drive} and DriveWorld~\cite{min2024driveworld} chain latent prediction with planning; LAW~\cite{li2025law} predicts future features self-supervisedly.
WoTE~\cite{li2025wote} uses a BEV world model with LiDAR--camera feature fusion for trajectory scoring.
Closest to ours is DINO-WM~\cite{garrido2024dinowm}, which freezes a DINOv2 encoder as the prediction target.
All these methods process dense token sets, so both backbone and predictor costs scale with image resolution.
We show that 75--90\% of tokens can be dropped before encoding with improved planning quality, indicating that fewer but better-targeted tokens reduce attention dilution in the downstream planner.

\paragraph{Token sparsification in vision transformers.}
DynamicViT~\cite{rao2021dynamicvit} and EViT~\cite{liang2022evit} prune tokens based on learned attention scores.
ToMe~\cite{bolya2023tome} merges similar tokens; ATS~\cite{fayyaz2022ats} learns adaptive sampling.
Register tokens~\cite{darcet2024registers} and Perceiver IO~\cite{jaegle2021perceiverio} compress information via cross-attention into a fixed-size bottleneck.
These approaches require encoding the full image first, because token importance depends on extracted features.
This creates a hard floor on backbone FLOPs.
LaTP~\cite{lu2024latp} uses LiDAR-derived distance information with a content-and-distance-aware indicator to prune tokens for an LVLM-based trajectory predictor.
While LaTP reduces downstream prefilling cost, it still computes all image embeddings before scoring.
Our selector uses raw geometric measurements (point counts, heights, motion residuals) and operates before the visual backbone.
Discarded patches never incur visual-encoding cost, enabling a genuinely sparse backbone.

\paragraph{LiDAR--camera fusion vs.\ visual compute routing.}
BEVFusion~\cite{liu2023bevfusion,liang2022bevfusion}, TransFuser~\cite{chitta2023transfuser}, and WoTE~\cite{li2025wote} fuse learned LiDAR and camera features for detection or planning.
Fusion enriches representation quality by merging complementary modalities; however, it does not reduce the dense visual encoding that dominates foundation-model world models.
We address a different bottleneck: where should the visual backbone spend its encoding budget?
In our design, LiDAR provides geometric routing signals and metric coordinates---patch selection indices plus 3D anchors---without a learned LiDAR feature tensor.
The vision encoder remains the sole representation backbone.
A controlled ablation (\cref{sec:ablation}) shows that adding a learned LiDAR-feature fusion branch does not improve planning, confirming that LiDAR's value here is geometric routing, not feature fusion.

\paragraph{End-to-end autonomous driving.}
UniAD~\cite{hu2023uniad} unifies perception, prediction, and planning.
VAD~\cite{jiang2024vad} uses vectorized representations.
SparseDrive~\cite{sun2024sparsedrive} employs sparse queries.
DiffusionDrive~\cite{liao2025diffusiondrive,liao2025diffusiondrivev2} plans via truncated diffusion.
These methods focus on task design or planning heads and are orthogonal to our contribution: efficient token routing that reduces upstream visual compute by an order of magnitude.
Direct comparison is non-trivial because they use different backbones, training data, and evaluation protocols.
On NAVSIM (\cref{tab:navsim}), where a standardized scoring function enables fair comparison, \gats surpasses the dense baseline by 3 EPDMS points.

\section{Method}
\label{sec:method}

\begin{figure*}[t]
  \centering
  \includegraphics[width=0.95\linewidth]{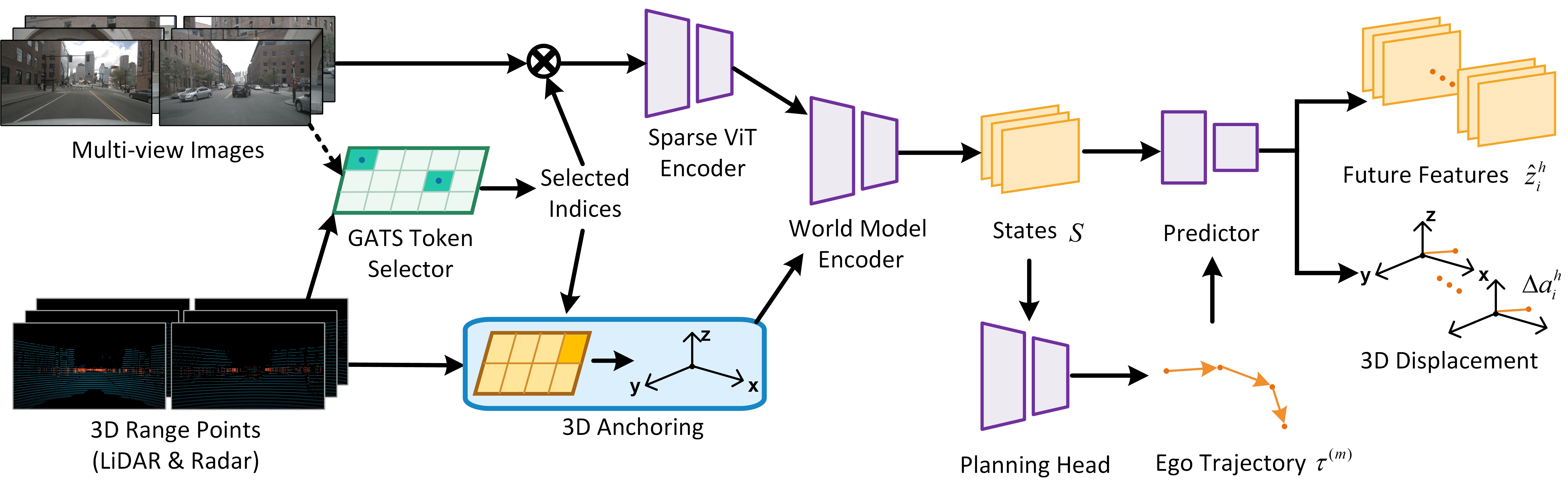}
  \vspace{-2mm}
  \caption{\textbf{Method overview.}
  LiDAR and radar measurements provide both patch-routing statistics and metric metadata.
  \gats selects image patches before visual encoding; Sparse ViT encodes only the selected patches.
  Their projected LiDAR anchors and geometric statistics are embedded as 3D positional metadata and added to the visual tokens before the world-model encoder.
  The same anchors provide metric references for displacement prediction and self-supervised feature-geometry matching.
  At 10\% budget the full pipeline costs 24 GFLOPs.}
  \label{fig:method}
  \vspace{-3mm}
\end{figure*}

\method has four stages (\cref{fig:method}): (1)~geometry-aware token selection (\gats) scores all image patches using raw LiDAR and radar measurements and outputs the \emph{indices} of the top-$K$ patches to retain; (2)~3D anchoring assigns each selected patch a metric coordinate in the current ego frame; (3)~a self-supervised latent world model predicts future visual features and 3D displacements; (4)~a planning head reads the encoded state to output an ego trajectory.
LiDAR and radar contribute geometric routing signals and metric coordinates---point-cloud projections and geometric statistics---not learned feature tensors.
No 3D backbone (e.g., PointPillars or VoxelNet) is required; the vision encoder remains the sole representation backbone.

\subsection{Geometry-Aware Token Selection}
\label{sec:gats}

Given $C$ cameras at resolution $H{\times}W$ with patch size~$p$, a single frame contains $N {=} C \cdot \frac{H}{p}\cdot\frac{W}{p}$ image tokens.
\gats retains $K = \lceil \beta N \rceil$ tokens per frame, where $\beta$ is the budget ratio (e.g., $\beta{=}0.1$).
Note that the world model ingests $T_h$ history frames, so the total token count entering the encoder is $T_h \cdot K$ plus ego-state tokens.

\paragraph{Geometric statistics.}
We project multi-sweep LiDAR points and radar detections onto each camera's image grid.
For patch~$i$ we aggregate a 16-dimensional geometry vector $\mathbf{g}_i$ encoding:
validity~$v_i{\in}\{0,1\}$ (1 if ${\geq}1$ LiDAR point falls in the patch, 0 otherwise; when $v_i{=}0$, all other entries are zeroed),
log point count~$\log(1{+}n_i)$,
depth statistics~($d_{\min,i}, d_{\text{med},i}, \sigma_{d,i}$),
height above the ground plane~($h_{\text{med},i}, h_{\max,i}, \sigma_{h,i}$),
non-ground fraction~$f_{\text{ng},i}$,
multi-sweep motion residuals~($r_{\mu,i}, r_{\max,i}$),
radar radial speed $|\bar{v}_{r,i}|$ and hit indicator~$\mathbf{1}_{r,i}$,
mean LiDAR intensity $I_i$, and 2D radar velocity~$(v_{rx,i}, v_{ry,i})$.

\paragraph{Three-stream scoring.}
The budget $K$ is allocated to three streams operating on disjoint candidate pools:
\begin{itemize}[leftmargin=*,topsep=2pt,itemsep=1pt]
  \item \textbf{Obstacle stream} (${\sim}55\%$ of $K$, non-ground patches with $f_{\text{ng},i}{>}0.3$): Scores by non-ground evidence, motion, and ego-path proximity.
  The allocation 55/30/15\% is a fixed hyperparameter tuned on the validation set; $f_{\text{ng},i}$ is the fraction of LiDAR points in the patch above the ground plane.
  \begin{equation}
    \begin{aligned}
    s_{\text{obj},i} &= 1.5\,f_{\text{ng},i} + 2.0\,\bar{r}_{\mu,i} + 2.0\,|\bar{v}_{r,i}| \\[-2pt]
    &\quad + e^{-d_{\text{med},i}/20} + p_i + \text{MLP}(\mathbf{g}_i)
    \end{aligned}
    \label{eq:sobj}
  \end{equation}
  where $\bar{\cdot}$ denotes clipping to $[0,1]$, $\bar{r}_{\mu,i}$ is the mean motion-residual magnitude, $e^{-d_{\text{med},i}/20}$ is a depth-decay term favouring closer objects (20\,m is a scale constant), $\text{MLP}$ is a 2-layer network (hidden dim 32, ReLU, trained end-to-end), and $p_i{=}\exp(-d_{\text{BEV},i}/5)$ is the ego-path proximity, where $d_{\text{BEV},i}$ is the BEV distance from the patch's LiDAR median to the extrapolated ego path (constant-turn-rate model from the current ego state).
  The fixed coefficients (1.5, 2.0, etc.) were tuned on the validation set; the learned MLP adapts them during training.
  \item \textbf{Road stream} (${\sim}30\%$, ground patches with $f_{\text{ng},i}{\le}0.3$):
  \begin{equation}
    s_{\text{road},i} = (1{-}f_{\text{ng},i}) + 0.5\,\sigma_{h,i} + 0.3\,\bar{I}_i + p_i + \text{MLP}_\text{road}(\mathbf{g}_i)  \notag
  \end{equation}
  targeting drivable-surface geometry and road reflectance cues near the ego corridor.
  \item \textbf{Visual reserve} (${\sim}15\%$, all patches): In dense-backbone training, uses DINOv2 [CLS] attention weights; in sparse-backbone deployment, falls back to a geometry-only heuristic based on far-range height cues.  This stream captures objects like elevated traffic signals that lack ground-level LiDAR returns.
\end{itemize}
Because the three streams draw from disjoint or complementary candidate pools, selected sets do not double-count; the union gives exactly $K$ tokens.
During training, each stream applies Gumbel noise for differentiable top-$K$ selection: the gate weight $w_i = 1 + \tilde{s}_i - \operatorname{sg}(\tilde{s}_i)$ (with $\tilde{s}_i {=} \operatorname{softmax}((s_i{+}\epsilon_i)/\tau)$ and $\operatorname{sg}$ the stop-gradient operator) multiplies the selected token features, providing a gradient path through the discrete selection.

\paragraph{Sparse backbone encoding.}
Because \gats operates only on 3D point-cloud projections, the ViT backbone encodes \emph{only} the $K$ selected patches across all cameras, prepending [CLS] and register tokens~\cite{darcet2024registers}.
Dense encoding costs ${\sim}$185 GFLOPs (DINOv2-Small, 6 cameras); sparse encoding at 10\% budget costs ${\sim}$16 GFLOPs.
The sparse backbone is distilled from the dense one (\cref{sec:efficiency}).

\subsection{3D-Anchored Tokens}
\label{sec:anchor}

Each selected token~$i$ receives a 3D anchor $\mathbf{a}_i \in \mathbb{R}^3$: the median LiDAR point position within its patch, expressed in the current ego frame.
For the visual-reserve patches that lack LiDAR points ($v_i{=}0$), we back-project along the camera ray at a default depth $d_0{=}30$\,m to produce a virtual anchor, flagged by $v_i$ so the model can distinguish measured from imputed coordinates.

The anchor enables:
\textbf{(i)}~cross-camera association via a shared 3D space;
\textbf{(ii)}~cross-frame alignment by transforming \emph{history} anchors $\mathbf{a}_i^{t-k}$ into the current ego frame via known odometry: $\tilde{\mathbf{a}}_i^{t-k} = \mathbf{T}_{t-k\to t}\,\mathbf{a}_i^{t-k}$;
\textbf{(iii)}~3D position encoding: $\mathrm{PE}_\text{3D} = \mathrm{MLP}\bigl([\gamma(\mathbf{a}_i \oslash \mathbf{s}),\;\sigma_{d,i}]\bigr)$, where $\gamma$ denotes sinusoidal Fourier features with 16 frequency bands, $\mathbf{s}=(50,50,4)^\top\text{m}$ is a normalization scale, and $\mathrm{PE}_\text{3D}$ is \emph{added} element-wise to the visual token embedding.

\subsection{Self-Supervised Latent World Model}
\label{sec:wm}

\paragraph{Encoder.}
The $K$ selected patches are first encoded by the Sparse ViT backbone into visual tokens.
These sparse visual tokens from $T_h$ history frames, augmented with ego-state tokens and per-camera context summaries, are then processed by the world-model encoder (a transformer) to produce state $\mathbf{S}$.
In dense-backbone mode, the context summary is a spatial average pool of the unselected patches; in sparse-backbone mode it is the ViT's [CLS] token.

\paragraph{Predictor.}
Given $\mathbf{S}$ and a future ego trajectory $\boldsymbol{\tau}_{1:H}$, a predictor transformer generates latent states $\mathbf{Z}^h$ for $H$ future steps.
Each input token $i$ produces one output per future step, maintaining a one-to-one correspondence.
Two heads decode per-token quantities:
\begin{itemize}[leftmargin=*,topsep=2pt,itemsep=1pt]
  \item \textbf{3D displacement} $\Delta\mathbf{a}_i^h \in \mathbb{R}^3$: the ego-frame motion of the anchor from time $t$ to $t{+}h\Delta t$.
  \item \textbf{Future feature} $\hat{\mathbf{z}}_i^h \in \mathbb{R}^{D_f}$: the predicted DINOv2 representation at the displaced position.
\end{itemize}

\paragraph{Self-supervised motion targets.}
Instead of box annotations, displacement ground truth comes from feature-geometry matching.
Each token at time $t$ is matched to its nearest counterpart at $t{+}\Delta t$ using a joint cost:
\begin{equation}
  j^* = \argmin_j \frac{\|\mathbf{f}_i^t - \mathbf{f}_j^{t+\Delta t}\|^2}{D_f} + \frac{\|\hat{\mathbf{a}}_i^t - \mathbf{a}_j^{t+\Delta t}\|^2}{r^2}
  \label{eq:match}
\end{equation}
where $\hat{\mathbf{a}}_i^t$ is a constant-velocity extrapolation from the previous match, $r{=}3$\,m is the match radius, and $\mathbf{a}_j^{t+\Delta t}$ is ego-motion-compensated into the frame at time $t$.
The pseudo-flow target is then $\Delta\mathbf{a}_i^{\text{gt}} = \mathbf{a}_{j^*}^{t+\Delta t} - \mathbf{a}_i^t$.
Ground-classified tokens are forced to $\Delta\mathbf{a}_i^{\text{gt}}{=}\mathbf{0}$ (static-world prior).

\paragraph{Training losses.}
\begin{itemize}[leftmargin=*,topsep=2pt,itemsep=1pt]
  \item \textbf{JEPA-style feature prediction loss:} MSE + cosine between $\hat{\mathbf{z}}_i^h$ and frozen DINOv2 targets.
  \item \textbf{Anchor loss:} The predicted future 3D position $\hat{\mathbf{a}}_i^h {=} \mathbf{a}_i + \Delta\mathbf{a}_i^h$ is projected back to the image plane via the camera intrinsics; the DINOv2 feature bilinearly sampled at that pixel \emph{from the frame at time $t{+}h\Delta t$} is compared with the predicted feature $\hat{\mathbf{z}}_i^h$.
  \item \textbf{Match loss:} Smooth-L1 between predicted and pseudo-flow displacements, up-weighted $5\times$ for tokens with $\|\Delta\mathbf{a}_i^{\text{gt}}\|{>}1$\,m.
  \item \textbf{Static regularization:} L2 penalty on $\|\Delta\mathbf{a}_i^h\|$ for ground-classified tokens.
\end{itemize}

\subsection{Planning Head}
\label{sec:plan}

A learnable query attends to the encoder state $\mathbf{S}$ via cross-attention and regresses a trajectory $\hat{\boldsymbol{\tau}} = (\hat{x}_t, \hat{y}_t, \hat{\psi}_t)_{t=1}^{T}$.
A vocabulary head classifies among $V{=}256$ trajectory prototypes; the top-$M$ candidates ($M{=}8$) are re-scored by the predictor: each candidate $\boldsymbol{\tau}^{(m)}$ is rolled out through the predictor to obtain future anchor positions $\hat{\mathbf{a}}_i^h(\boldsymbol{\tau}^{(m)})$, and the candidate that minimizes the predicted collision cost is selected.
The collision cost counts the number of predicted non-ground anchors within a $2{\times}5$\,m ego bounding box (enlarged for safety margin) placed along each candidate trajectory; the final output is the regression trajectory $\hat{\boldsymbol{\tau}}$ when the minimum-cost candidate agrees with the regression, and the vocabulary candidate otherwise.

\section{Experiments}
\label{sec:exp}

We organize experiments around four questions:
\textbf{(Q1)}~Does \gats beat dense models in planning quality? (\cref{sec:main_results})
\textbf{(Q2)}~Does it cover safety-critical objects? (\cref{sec:intent})
\textbf{(Q3)}~Can the world model learn motion without box labels? (\cref{sec:motion})
\textbf{(Q4)}~How far can end-to-end FLOPs be reduced? (\cref{sec:efficiency})
Ablation (\cref{sec:ablation}) and robustness (\cref{sec:robust}) studies complete the picture.

\subsection{Setup}
\label{sec:setup}

\paragraph{Datasets.}
We evaluate on nuScenes~\cite{caesar2020nuscenes} (28k train / 6k val keyframes, 6 cameras, LiDAR, radar) and NAVSIM~\cite{dauner2024navsim} navtest (8 cameras, LiDAR; 85k train / 12k test with the official scoring function).
NAVSIM does not provide radar; the radar stream is zeroed for NAVSIM experiments. Detailed definitions of all evaluation metrics are provided in \cref{app:metrics}.

\paragraph{Architecture.}
The vision backbone is DINOv2-Small with registers~\cite{oquab2024dinov2,darcet2024registers} ($D{=}384$, 12 layers), encoding images at $252{\times}448$ with patch size~14, yielding $P{=}576$ patches per camera and $N{=}3{,}456$ tokens per frame (nuScenes, 6 cameras) or $N{=}4{,}608$ (NAVSIM, 8 cameras).
The world model uses $d{=}256$, 4 world-model encoder layers, 3 predictor layers, 2 feature-readout layers, and 2 planning layers (${\sim}$10M parameters).
History depth $T_h{=}2$; prediction horizon $H{=}4$ at 0.5\,s steps.

\paragraph{Collision rate.}
We place the ego bounding box ($4.1{\times}1.9$\,m) along the predicted trajectory at each of the 6 future time steps (3\,s) and check geometric overlap with GT 3D boxes.
Following UniAD~\cite{hu2023uniad}, we exclude timesteps where the GT ego trajectory already collides with GT boxes (annotation noise), so the metric reflects only planner-induced collisions.

\paragraph{Training.}
All models train for 12 epochs on cached DINOv2 features with AdamW ($\text{lr}{=}4{\times}10^{-4}$, cosine decay, batch size~16).
Total training time is ${\sim}$40 minutes per configuration on a single RTX 4090.
Sparse backbone distillation adds 6 epochs.

\paragraph{Baselines.}
We compare \gats against:
\textbf{Dense fusion} (all tokens, LiDAR anchoring);
\textbf{Dense camera} (all tokens, no LiDAR);
\textbf{Random};
\textbf{Grid} (uniform spatial);
\textbf{DINOv2-Attn} (DINOv2 [CLS] attention);
\textbf{Learned}~\cite{jaegle2021perceiverio};
\textbf{Distance} (LiDAR-depth heuristic, cf.~\cite{lu2024latp});
\textbf{Register}~\cite{darcet2024registers};
\textbf{Pool} (adaptive avg.\ pooling).
All sparse methods share the same world model and 3D anchoring.
\textbf{Note:} the Distance baseline selects patches by inverse LiDAR depth---a heuristic inspired by LaTP~\cite{lu2024latp}.
It does not reproduce the full LaTP pipeline; the comparison is at the selection-principle level.

\subsection{Main Results (Q1: Planning Quality)}
\label{sec:main_results}

\begin{table}[t]
\centering
\caption{\textbf{Planning quality vs.\ compute on nuScenes val.}
L2: avg.\ displacement error (m) over 3\,s.
Col: collision rate (\%, excluding GT-collision timesteps, following~\cite{hu2023uniad}).
GFLOPs: world model + planner only.}
\label{tab:main_nusc}
\vspace{-3mm}
\resizebox{\columnwidth}{!}{%
\begin{tabular}{lccccc}
\toprule
Method & $\beta$ & Tok. & L2$_\text{avg}$$\downarrow$ & Col\%$\downarrow$ & GFLOPs \\
\midrule
\multicolumn{6}{l}{\textit{Published E2E (box+map sup., different backbone):}} \\
UniAD~\cite{hu2023uniad} & 100\% & --- & 1.02$^\ddag$ & 0.29$^\ddag$ & --- \\
VAD-Base~\cite{jiang2024vad} & 100\% & --- & 0.54$^\ddag$ & 0.07$^\ddag$ & --- \\
SparseDrive~\cite{sun2024sparsedrive} & 100\% & --- & 0.58$^\ddag$ & 0.05$^\ddag$ & --- \\
\midrule
\multicolumn{6}{l}{\textit{Our models (DINOv2-S, self-supervised, no box/map ann.):}} \\
Dense fusion & 100\% & 6913 & 0.573 & 0.09 & 248 \\
Dense camera & 100\% & 6913 & 0.668 & 0.16 & 245 \\
\midrule
GATS & 10\% & 717 & 0.537 & 0.09 & 7.6 \\
\textbf{GATS} & \textbf{25\%} & \textbf{1753} & \textbf{0.520} & \textbf{0.04} & \textbf{25.9} \\
\midrule
Random & 25\% & 1753 & 0.550 & 0.09 & 25.7 \\
Grid & 25\% & 1753 & 0.604 & 0.12 & 25.7 \\
DINOv2-Attn & 25\% & 1753 & 0.584 & 0.09 & 25.7 \\
Distance (depth)~\cite{lu2024latp} & 25\% & 1753 & 0.631 & 0.23 & 25.7 \\
Learned~\cite{jaegle2021perceiverio} & 25\% & 1753 & 0.572 & 0.19 & 26.4 \\
Register~\cite{darcet2024registers} & 25\% & 1753 & 0.566 & 0.16 & 40.8 \\
Pool & 25\% & 1761 & 0.595 & 0.13 & 25.7 \\
\bottomrule
\end{tabular}%
}
\begin{flushleft}
\footnotesize{$^\ddag$Published numbers use different backbones (ResNet-50/101) and full 3D box / HD-map supervision.}
\end{flushleft}
\vspace{-4mm}
\end{table}

\Cref{tab:main_nusc} answers Q1.
\gats at 25\% budget (1{,}753 tokens) achieves the best planning L2 in the table: 0.520\,m---outperforming the dense-fusion baseline (0.573\,m, 6{,}913 tokens) by 9\% while using $10{\times}$ fewer world-model FLOPs.
Random selection at 25\% (0.550\,m) already outperforms the dense model (0.573\,m).
This confirms the attention-dilution hypothesis (\cref{tab:entropy}): dense tokens overwhelm the planner's cross-attention, and removing irrelevant patches is more valuable than retaining them.
\gats further improves over random by 0.030\,m via geometry-informed routing.

The results reveal two practical operating points:
\emph{(i)}~25\% budget for maximum quality ($10{\times}$ WM speedup, $6.6{\times}$ end-to-end);
\emph{(ii)}~10\% budget for extreme efficiency (${\sim}33{\times}$ WM speedup, $18{\times}$ end-to-end) while still beating dense on L2 (0.537\,m vs.\ 0.573\,m).

\begin{figure}[t]
  \centering
  \includegraphics[width=\columnwidth,trim=0 5 0 5,clip]{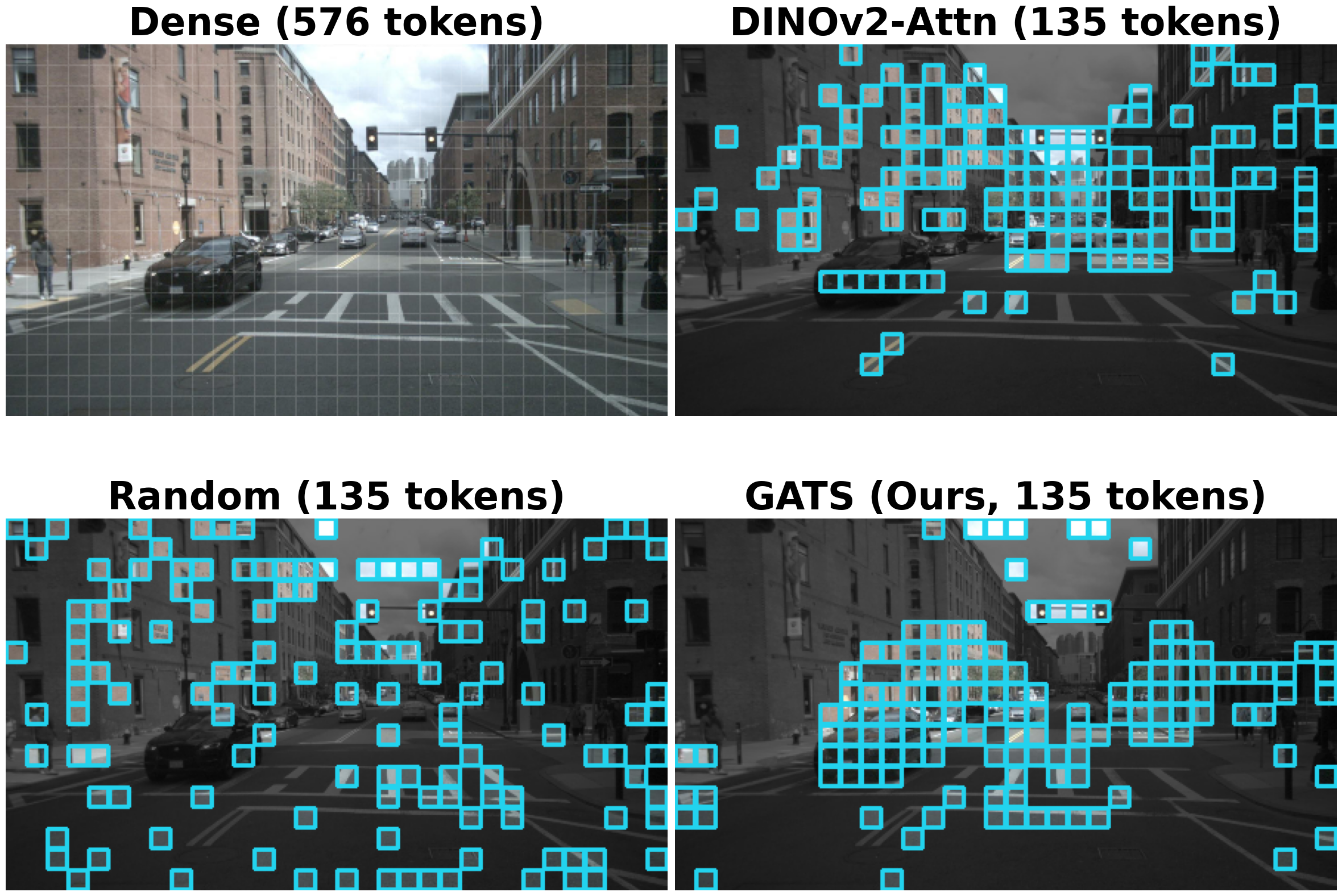}
  \vspace{-6mm}
  \caption{\textbf{Token selection comparison} (front camera, 25\% global budget). \emph{Dense:} all 576 tokens. For controlled visual comparison, Random and DINOv2-Attn are shown with the same front-camera token count as \gats at this scene. \gats concentrates on road users and drivable surface near the ego path.}
  \label{fig:comparison}
  \vspace{-5mm}
\end{figure}

\begin{figure*}[t]
  \centering
  \includegraphics[width=0.80\linewidth,trim=0 5 0 23,clip]{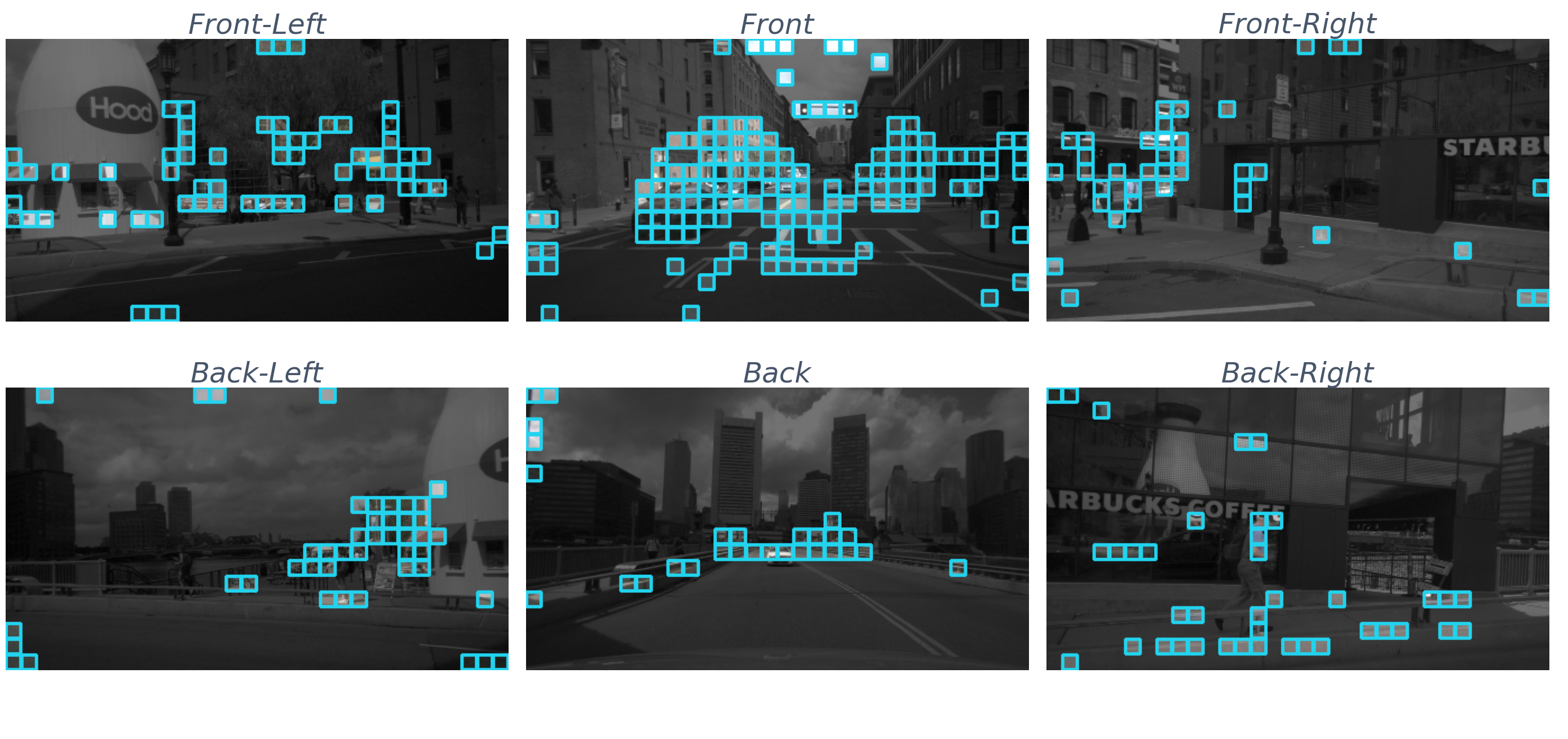}
  \vspace{-4mm}
  \caption{\textbf{Qualitative result.}
  Six camera views with \gats selection at 10\% budget; bright patches are retained tokens, dark regions are pruned.
  Most driving-relevant structure is covered while sky and irrelevant background elements remain pruned.}
  \label{fig:qual}
  \vspace{-4mm}

\end{figure*}

\Cref{fig:comparison} compares selection strategies on the front camera; \cref{fig:qual} shows all six surround cameras at 10\% budget, where the contrast between kept and pruned regions is most apparent.
The intent-critical analysis (\cref{sec:intent}) provides a more discriminative safety comparison.

\vspace{-1mm}
\paragraph{NAVSIM benchmark.}

\begin{table}[t]
\centering
\small
\caption{\textbf{NAVSIM v2 navtest results (EPDMS).}
All our models use the same codebase and v2 scoring function.
GFLOPs: world model + planner only (NAVSIM uses 8 cameras, hence higher than nuScenes).}
\label{tab:navsim}
\vspace{-3mm}
\resizebox{\columnwidth}{!}{%
\begin{tabular}{lccc}
\toprule
Method & $\beta$ & EPDMS$\uparrow$ & GFLOPs \\
\midrule
\multicolumn{4}{l}{\textit{Published (different backbone / supervision):}} \\
WoTE~\cite{li2025wote} & 100\% & 87.7 & --- \\
DiffusionDrive~\cite{liao2025diffusiondrive} & 100\% & 85.2 & --- \\
DiffusionDrive-v2~\cite{liao2025diffusiondrivev2} & 100\% & 85.5 & --- \\
\midrule
\multicolumn{4}{l}{\textit{Our models (DINOv2-S, self-supervised):}} \\
Dense fusion & 100\% & 85.9 & 722 \\
Dense camera & 100\% & 85.5 & 719 \\
\textbf{GATS} & \textbf{25\%} & \textbf{88.9} & \textbf{68} \\
GATS & 10\% & 87.9 & 18 \\
Random & 25\% & 85.2 & 68 \\
Grid & 25\% & 85.4 & 68 \\
DINOv2-Attn & 25\% & 85.1 & 68 \\
Distance (depth) & 25\% & 81.7 & 68 \\
Learned~\cite{jaegle2021perceiverio} & 25\% & 84.8 & 68 \\
Register~\cite{darcet2024registers} & 25\% & 85.3 & 68 \\
Pool & 25\% & 86.7 & 68 \\
\bottomrule
\end{tabular}%
}
\begin{flushleft}
\end{flushleft}
\vspace{-4mm}
\end{table}

On NAVSIM v2 (\cref{tab:navsim}), \gats at 25\% achieves an EPDMS of 88.9---surpassing both dense fusion (85.9) and all sparse baselines (81.7--86.7), as well as published methods WoTE (87.7), DiffusionDrive (85.2), and its RL-enhanced successor DiffusionDrive-v2 (85.5).
At 10\%, \gats (87.9 EPDMS) still exceeds all published baselines while using ${\sim}40{\times}$ fewer world-model FLOPs than our own dense model.
The published methods address a different design axis (planning-head architecture, trajectory diffusion) and use different backbones and supervision; the comparison serves as a reference point rather than a controlled ablation.

\vspace{-1mm}
\subsection{Safety-Critical Coverage (Q2)}
\label{sec:intent}

Average planning scores are dominated by uneventful scenes.
We mine 739 intent-critical samples by thresholding GT ego deceleration ($>$2\,m/s$^2$) or lateral deviation ($>$0.5\,m) within the 3\,s horizon, indicating hard braking or evasive maneuvers.
A \emph{hazardous vehicle} is the GT vehicle whose 3D box is closest to any future ego waypoint in that sample (i.e., the agent most likely to cause a collision).
For each sample, we ask: does the selector place at least one token on the hazardous vehicle?

\begin{table}[t]
\centering
\small
\caption{\textbf{Intent-critical analysis} on 739 mined samples at 25\% budget.
Cov.: hazardous vehicles with $\geq$1 selected token. Col.: collision rate.}
\label{tab:intent}
\vspace{-3mm}
\begin{tabular*}{\columnwidth}{@{\extracolsep{\fill}}lcc}
\toprule
Selector ($\beta{=}25\%$) & Cov.$\uparrow$ & Col.$\downarrow$ \\
\midrule
\textbf{GATS} & \textbf{95.8\%} & \textbf{0.09\%} \\
DINOv2-Attn & 93.9\% & 0.13\% \\
Random & 81.5\% & 0.25\% \\
Grid & 80.5\% & 0.43\% \\
Learned~\cite{jaegle2021perceiverio} & 70.0\% & 0.61\% \\
Distance (depth)~\cite{lu2024latp} & 22.1\% & 0.50\% \\
\midrule
Dense fusion & 96.8\% & 0.34\% \\
\bottomrule
\end{tabular*}
\vspace{-4mm}
\end{table}

\Cref{tab:intent} reveals two findings.
First, \gats covers 95.8\% of hazardous vehicles at 25\% budget---approaching the 96.8\% ceiling of the dense model and far above random (81.5\%), grid (80.5\%), and learned/distance selectors (70.0\%/22.1\%).
DINOv2-Attn reaches similar coverage (93.9\%) thanks to its CLS-attention mechanism, but achieves worse planning quality overall (\cref{tab:main_nusc}).

Second, \gats achieves a lower collision rate (0.09\%) on these hard samples than the dense model (0.34\%), despite processing only 25\% of the tokens.
We attribute this to attention concentration: the dense model feeds 6{,}913 tokens to the planner's cross-attention, of which a typical vehicle occupies only 3 tokens (median; 7 at the 75th percentile).
The planning signal is diluted by thousands of irrelevant background tokens.
Direct measurement confirms this: the planner's cross-attention entropy is 7.0 bits for dense inputs vs.\ 5.7 bits for \gats (6.5 for random), as shown in \cref{tab:entropy}---\gats concentrates attention on fewer, more relevant tokens, yielding more decisive avoidance maneuvers.

\begin{table}[t]
\centering
\small
\caption{\textbf{Planner cross-attention entropy} (bits, lower = more concentrated). Measured over 800 nuScenes val samples. $N_\text{tok}$: per-frame token count (the world model ingests $T_h{=}2$ frames, but entropy is measured per frame). Max.\ possible entropy: $\log_2 N_\text{tok}$.}
\label{tab:entropy}
\vspace{-3mm}
\begin{tabular*}{\columnwidth}{@{\extracolsep{\fill}}lccc}
\toprule
Input & $N_\text{tok}$ & Entropy$\downarrow$ & $\log_2 N_\text{tok}$ \\
\midrule
Dense fusion & 3{,}456 & 7.0 & 11.8 \\
Random (25\%) & 864 & 6.5 & 9.8 \\
\textbf{GATS (25\%)} & \textbf{864} & \textbf{5.7} & 9.8 \\
\bottomrule
\end{tabular*}
\vspace{-3mm}
\end{table}

\vspace{-1mm}
\subsection{Motion Prediction (Q3)}
\label{sec:motion}

\begin{table}[t]
\centering
\small
\caption{\textbf{Motion prediction on nuScenes val} (EPE in m at 2\,s, no box supervision).
Zero@2s is the mean GT travel distance of each selector's moving tokens (higher = harder targets).
Rel.\% measures reduction from this zero-motion baseline.
EPE$_\text{mov}$ is not directly comparable across methods because each selects different points.}
\label{tab:motion}
\vspace{-3mm}
\resizebox{\columnwidth}{!}{%
\begin{tabular}{lcccc}
\toprule
Method & EPE$_\text{mov}$@2s$\downarrow$ & Zero@2s$\uparrow$ & Rel.\%$\uparrow$ & EPE$_\text{sta}$@2s$\downarrow$ \\
\midrule
\textbf{GATS (25\%)} & \textbf{5.76} & \textbf{12.94} & \textbf{55.5\%} & \textbf{0.54} \\
Random (25\%) & 6.81 & 11.93 & 42.9\% & 0.58 \\
DINOv2-Attn (25\%) & 6.65 & 11.55 & 42.4\% & 0.56 \\
Dense fusion & 6.62 & 11.92 & 44.5\% & 0.59 \\
\bottomrule
\end{tabular}%
}
\vspace{-3mm}
\end{table}

\Cref{tab:motion} evaluates the world model's ability to predict unsupervised motion.
A key subtlety: Zero@2s is the zero-motion baseline---the error if all selected moving tokens are predicted to stay still.
Because \gats specifically targets fast-moving objects (via motion residuals and radar), its Zero@2s (12.94\,m) is higher than DINOv2-Attn's (11.55\,m), meaning \gats is tested on harder, faster-moving targets.

The Rel.\% column normalizes for this difficulty difference: \gats reduces the zero-baseline error by 55.5\%, substantially above DINOv2-Attn (42.4\%) and random (42.9\%).
Static objects achieve sub-meter accuracy across all methods (0.54--0.59\,m), validating the ground-point static prior.

\vspace{-1mm}
\subsection{Ablation Study}
\label{sec:ablation}

\begin{table}[t]
\centering
\small
\caption{\textbf{Ablation} on nuScenes val (baseline: GATS 25\%).
Each row removes one component; $\Delta_\text{L2}$ is the relative planning degradation.}
\label{tab:ablation}
\vspace{-3mm}
\begin{tabular*}{\columnwidth}{@{\extracolsep{\fill}}lccc}
\toprule
Variant & L2$_\text{avg}$$\downarrow$ & $\Delta_\text{L2}$ & Col\%$\downarrow$ \\
\midrule
Full (GATS 25\%) & \textbf{0.520} & -- & \textbf{0.04} \\
\midrule
w/o 3D anchoring & 0.958 & +84.2\% & 0.43 \\
w/o road stream & 0.609 & +17.1\% & 0.13 \\
w/o learned refine. & 0.606 & +16.5\% & 0.09 \\
w/o world model & 0.574 & +10.4\% & 0.14 \\
w/o past flow & 0.569 & +9.4\% & 0.11 \\
w/o radar velocity & 0.561 & +7.9\% & 0.16 \\
w/o motion residuals & 0.540 & +3.8\% & 0.09 \\
w/o static prior & 0.534 & +2.7\% & 0.08 \\
\bottomrule
\end{tabular*}
\vspace{-3mm}
\end{table}

\Cref{tab:ablation} shows each component contributes meaningfully at 25\% budget.
Removing 3D anchoring is catastrophic (+84.2\%), confirming that metric coordinates are essential for cross-camera and temporal coherence in the sparse regime.
The road stream and learned refinement are the most impactful selection components (+17.1\% and +16.5\%): without the road stream, the selector concentrates on obstacles only and misses drivable-surface context; without learned refinement, the measured-geometry prior cannot adapt during training.
The world model contributes +10.4\%, confirming that self-supervised prediction shapes the encoder beyond auxiliary regularization.
Past flow (+9.4\%) and radar velocity (+7.9\%) validate the importance of temporal cues.
The same ablation on NAVSIM appears in \cref{app:navsim_ablation}.

\paragraph{Does deep LiDAR-feature fusion help?}
A natural question is why we do not fuse learned LiDAR features into the planning representation.
We add a 4-layer MLP (2.4M parameters) that maps the 16-dim geometry vector into a per-token embedding, fused additively after the ViT encoder.
At 25\% budget, this model achieves L2 = 0.569\,m---worse than \gats without fusion (0.520\,m), despite using 24\% more parameters.
At 10\%, the gap narrows (0.540\,m vs.\ 0.537\,m L2) but fusion still does not help.
We interpret this as further evidence for the attention-dilution hypothesis: additional features do not compensate for noisy attention over irrelevant tokens.
LiDAR's value in our pipeline is geometric routing, not feature fusion.

\vspace{-1mm}
\subsection{Robustness}
\label{sec:robust}

\begin{table}[t]
\centering
\small
\caption{\textbf{Robustness to image corruption} (severity 3, 25\% budget).
Clean L2 and corrupted L2 (avg.\ over 5 types: Gaussian noise, shot noise, motion blur, defocus blur, brightness) in meters.}
\label{tab:robust}
\vspace{-3mm}
\resizebox{\columnwidth}{!}{%
\begin{tabular}{lcccc}
\toprule
Method & Clean L2$\downarrow$ & Corrupt L2$\downarrow$ & Rel.\ degrad.$\downarrow$ & Mot.\ blur$\downarrow$ \\
\midrule
Dense camera & 0.668 & 0.924 & +38.3\% & +130\% \\
Dense fusion & 0.573 & 0.723 & +26.2\% & +88\% \\
\midrule
\textbf{GATS (25\%)} & \textbf{0.520} & \textbf{0.597} & +14.8\% & +36\% \\
Random (25\%) & 0.550 & 0.599 & +8.8\% & +20\% \\
Grid (25\%) & 0.604 & 0.678 & +12.2\% & +32\% \\
DINOv2-Attn (25\%) & 0.584 & 0.690 & +18.0\% & +44\% \\
\bottomrule
\end{tabular}%
}
\vspace{-2mm}
\end{table}

\Cref{tab:robust} shows both clean and corrupted absolute L2 errors at 25\% budget.
\gats has the best clean L2 (0.520\,m) and best corrupted L2 (0.597\,m) among all methods, surpassing random (0.599), dense fusion (0.723), and dense camera (0.924).
Sparse methods degrade 8.8--18\% under corruption, compared with 26--38\% for dense models, because their selection does not depend on pixel quality.
DINOv2-Attn degrades more (+18\%) than other sparse methods because its CLS-attention scoring is directly affected by corrupted pixels.
Under motion blur specifically, dense models degrade 88--130\% while \gats degrades only 36\%.

\vspace{-1mm}
\subsection{End-to-End Efficiency (Q4)}
\label{sec:efficiency}

\begin{table}[t]
\centering
\small
\caption{\textbf{End-to-end FLOPs (GFLOPs).} BB: vision backbone; WM: world model + planner.}
\label{tab:flops}
\vspace{-3mm}
\begin{tabular*}{\columnwidth}{@{\extracolsep{\fill}}lccc}
\toprule
Configuration & BB & WM & Total \\
\midrule
Dense (6 cams) & 185 & 248 & 433 \\
\textbf{Sparse 25\% (sparse BB)} & \textbf{40} & \textbf{25.9} & \textbf{66} \\
Sparse 10\% (sparse BB) & 16 & 7.6 & 24 \\
\bottomrule
\end{tabular*}
\begin{flushleft}
\footnotesize{25\%: $6.6\times$ total reduction. 10\%: $18\times$ total reduction. Sparse BB cosine sim.\ vs.\ dense: 0.87. Selector cost ($<$0.01 GFLOPs: point projection + MLP scoring) is included in BB.}
\end{flushleft}
\vspace{-6mm}
\end{table}

\Cref{tab:flops} shows two practical operating points.
\textbf{At 25\% budget}, the full pipeline costs 66 GFLOPs ($6.6\times$ reduction), with planning quality improving to 0.520\,m L2 (nuScenes) and an EPDMS of 88.9 (NAVSIM), both surpassing the dense baseline.
\textbf{At 10\% budget}, total cost drops to 24 GFLOPs ($18\times$ reduction); L2 = 0.537\,m still beats dense (0.573\,m) and EPDMS = 87.9 surpasses dense (85.9).
The sparse backbone is distilled in two phases (3 + 3 epochs); cosine similarity with the dense encoder recovers to 0.87.

\section{Conclusion}
\label{sec:conclusion}

A common assumption in vision-centric driving systems is that richer, denser representations lead to better planning.
Our results suggest otherwise.
By measuring cross-attention entropy, we show that dense token sets dilute the planner's focus, and that geometry-guided sparsity concentrates attention where it matters.
\method selects 10\% or 25\% of image tokens using LiDAR and radar geometry before the vision backbone runs, achieving 0.520\,m L2 on nuScenes and 88.9 EPDMS on NAVSIM---surpassing the dense baseline at $6.6\times$ fewer end-to-end FLOPs.
The finding that fewer, better-targeted tokens outperform exhaustive encoding suggests a broader principle: for task-specific inference, concentrated representations may be more valuable than comprehensive ones.

\paragraph{Limitations}
Patch resolution ($14{\times}14$ pixels) is too coarse for fine-grained cues such as turn signals; higher-resolution encoding of selected tokens is a natural extension.
The method requires LiDAR, which limits applicability to camera-only vehicles.
We report FLOPs as a hardware-agnostic compute proxy; real-time latency depends on implementation and hardware.

\paragraph{Broader impact.}
Efficient, safety-aware perception is important for deploying autonomous vehicles at scale.
By cutting compute by an order of magnitude while maintaining hazard coverage, \method makes latent world models more practical for real-time on-vehicle inference.
The principle that concentrated, task-relevant representations can outperform dense exhaustive encoding may also inform efficient inference in medical imaging, satellite analysis, or robotic manipulation.

{
    \small
    \bibliographystyle{ieeenat_fullname}
    \bibliography{main}
}

\maketitlesupplementary
\appendix

\renewcommand\thesection{A}
\renewcommand\theHsection{A}
\renewcommand\thetable{A\arabic{table}}
\renewcommand\theHtable{A\arabic{table}}
\renewcommand\thefigure{A\arabic{figure}}
\renewcommand\theHfigure{A\arabic{figure}}
\setcounter{table}{0}
\setcounter{figure}{0}
\setcounter{subsection}{0}

We provide detailed metric definitions, additional qualitative results, backbone analysis, NAVSIM ablation, and full 10\%\ budget tables. All GATS results in this paper are evaluated end-to-end with the distilled sparse backbone in deployment mode.

\subsection{Detailed Evaluation Metrics}
\label{app:metrics}

For completeness, we provide detailed definitions of all evaluation metrics used in the main paper:

\begin{enumerate}
  \item \textbf{3-sec displacement error ($\text{L2}_\text{avg}$):} On nuScenes, we evaluate open-loop planning over a 3\,s horizon (6 future timesteps at 0.5\,s intervals). $\text{L2}_\text{avg}$ measures the average 2D Euclidean distance (in meters) between the predicted ego trajectory waypoints and the ground-truth waypoints.
  \item \textbf{Collision rate (Col\%):} We place the ego bounding box ($4.1{\times}1.9$\,m) along the predicted trajectory at each of the 6 future time steps (3\,s) and check geometric overlap with GT 3D boxes.
Following UniAD~\cite{hu2023uniad}, we exclude timesteps where the GT ego trajectory already collides with GT boxes (annotation noise), so the metric reflects only planner-induced collisions. 
  \item \textbf{Extended Predictive Driver Model Score (EPDMS):} On NAVSIM v2~\cite{dauner2024navsim}, we report the official EPDMS, a composite non-reactive simulation metric in $[0, 100]$ that aggregates sub-scores for no at-fault collisions, drivable area compliance, driving direction compliance, time-to-collision (TTC), ego progress, and comfort.
  \item \textbf{Hazardous vehicle coverage (Cov.\%):} Evaluated on 739 mined intent-critical samples (\cref{sec:intent}). It measures the percentage of hazardous vehicles that have at least one selected token placed on them by the token selector.
  \item \textbf{Cross-attention entropy:} Measured in bits on nuScenes validation samples (\cref{tab:entropy}). It quantifies how concentrated or uniform the planner's attention is across the input tokens, where lower entropy indicates a more focused planning signal.
  \item \textbf{2-s End-Point Error ($\text{EPE}@2\text{s}$):} Evaluates unsupervised motion prediction at a 2-second horizon (in meters; \cref{sec:motion}). It is calculated as the 3D Euclidean distance between predicted and pseudo-ground-truth token displacements, reported separately for moving ($\text{EPE}_\text{mov}$) and static ($\text{EPE}_\text{sta}$) tokens.
\end{enumerate}

\subsection{25\% Budget Qualitative Results}
\label{app:qual_25}

\begin{figure*}[t]
  \centering
  \includegraphics[width=0.95\linewidth]{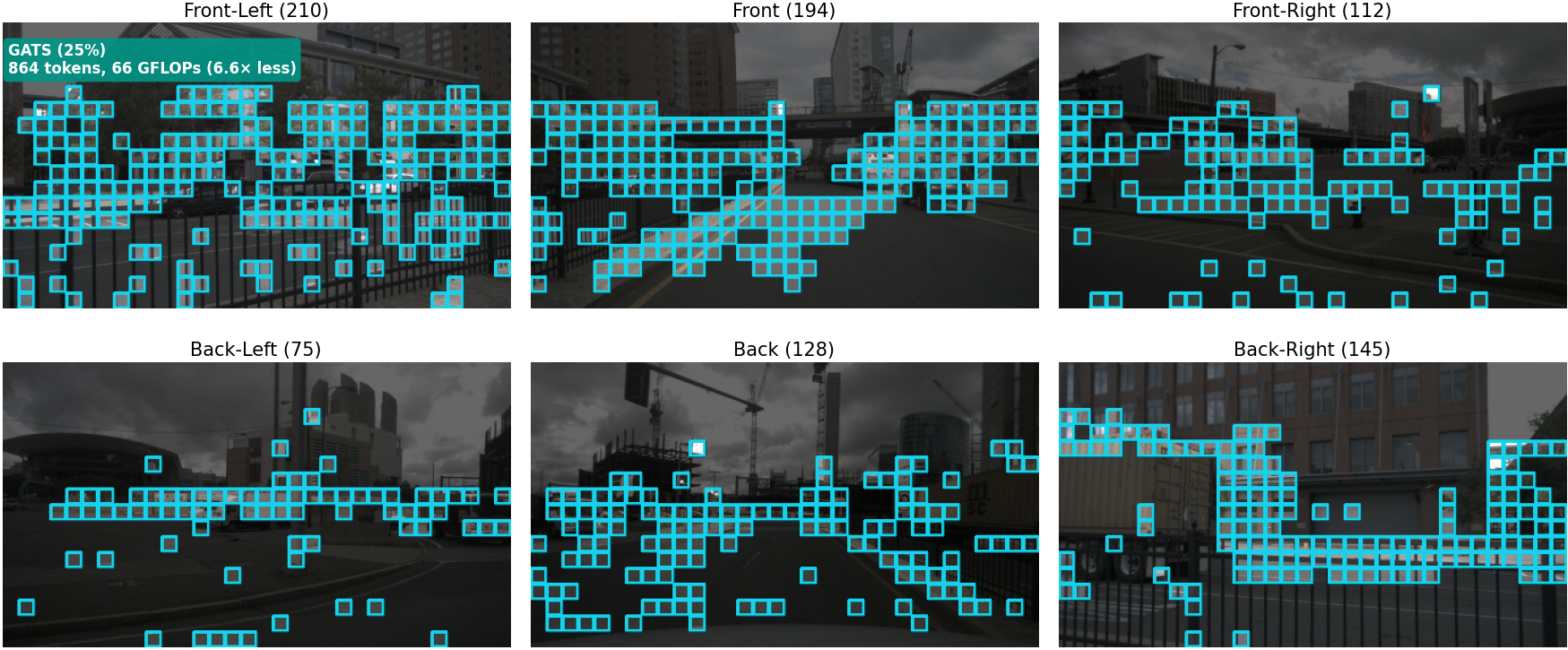}\\[12pt]
  \includegraphics[width=0.95\linewidth]{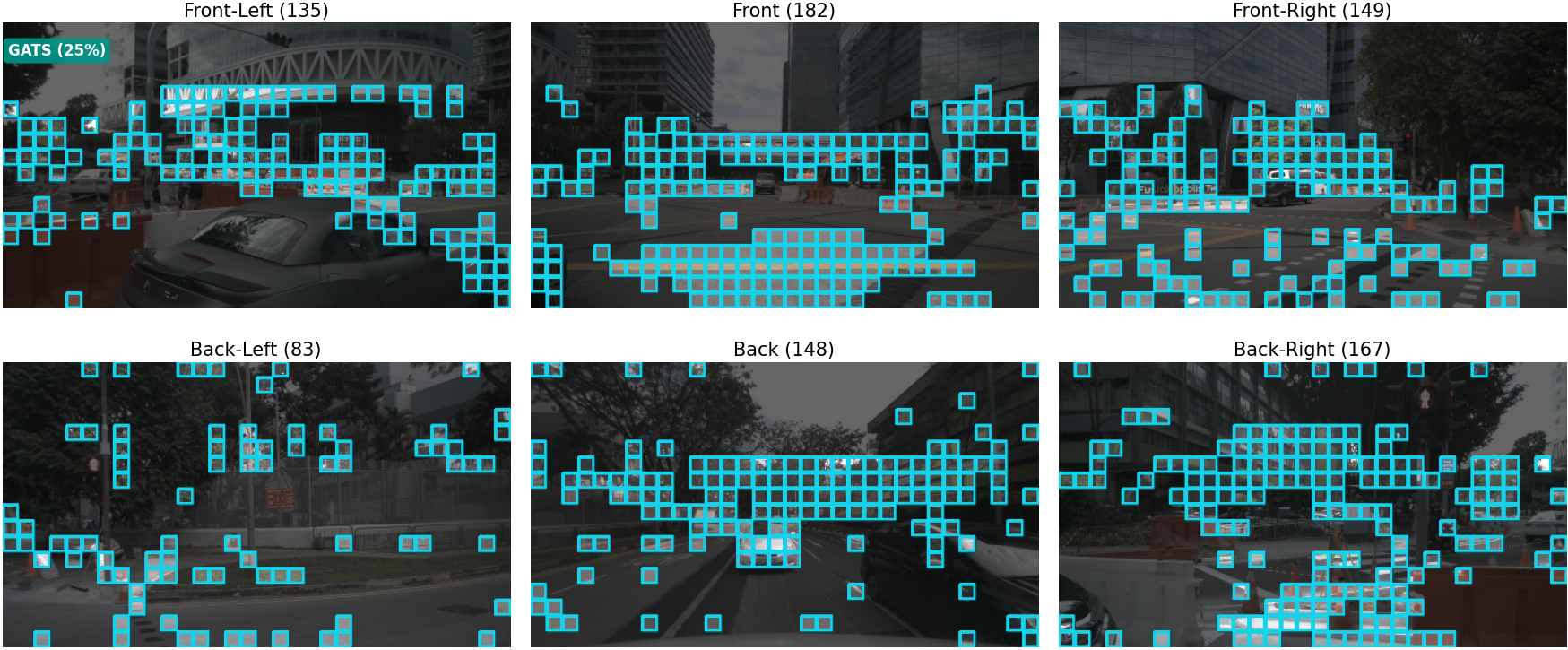}\\[12pt]
  \includegraphics[width=0.95\linewidth]{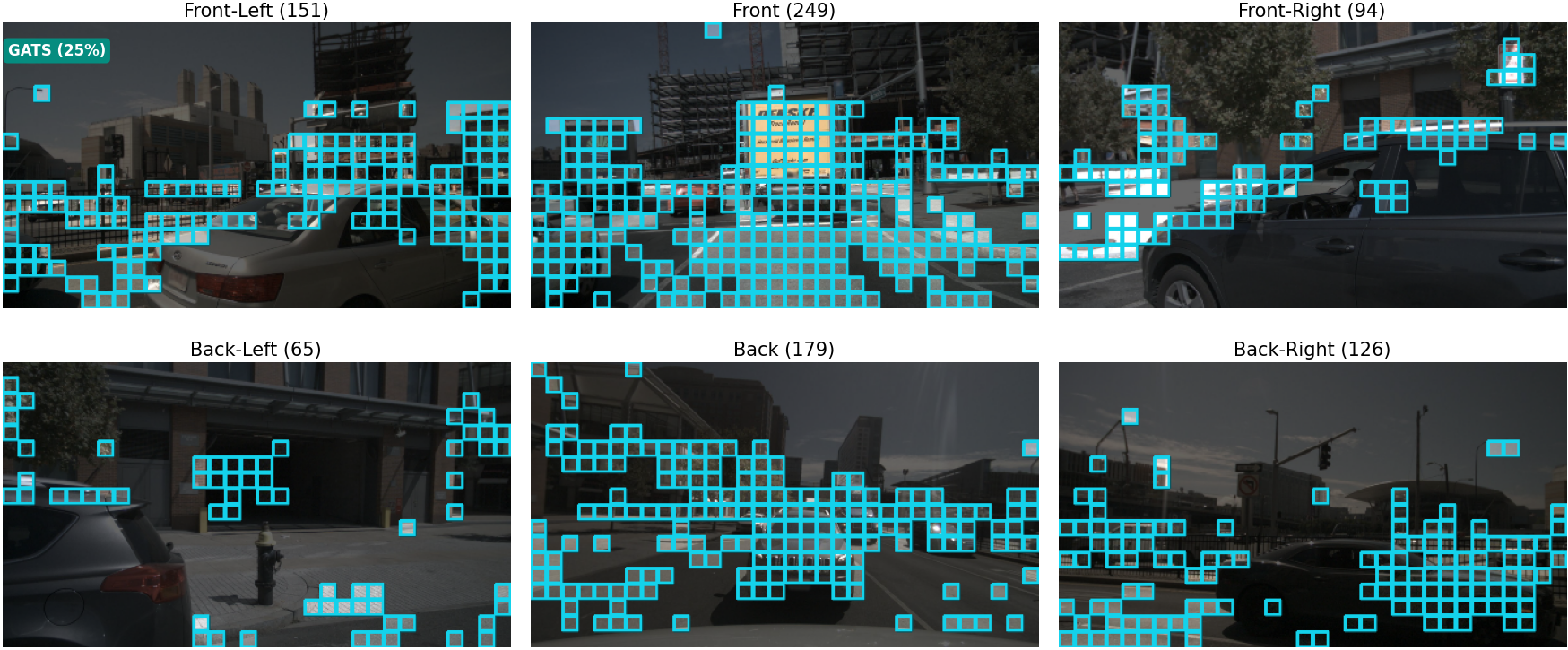}
  \caption{\textbf{\gats at 25\% budget} on three intent-critical scenes (6 surround cameras).
  Per-camera token counts (in parentheses) vary because \gats allocates globally: cameras facing busy road areas receive more tokens.
  At 25\%, most driving-relevant structure is covered while sky and distant buildings remain pruned.}
  \label{fig:appx_25pct}
\end{figure*}

At 25\% budget (864 tokens, 66 GFLOPs), \gats selects a denser set of patches while maintaining the same spatial concentration pattern as at 10\%.
\Cref{fig:appx_25pct} shows three representative scenes.

\subsection{10\% Budget Qualitative Results}
\label{app:qual_10}

\begin{figure*}[t]
  \centering
  \includegraphics[width=0.95\linewidth]{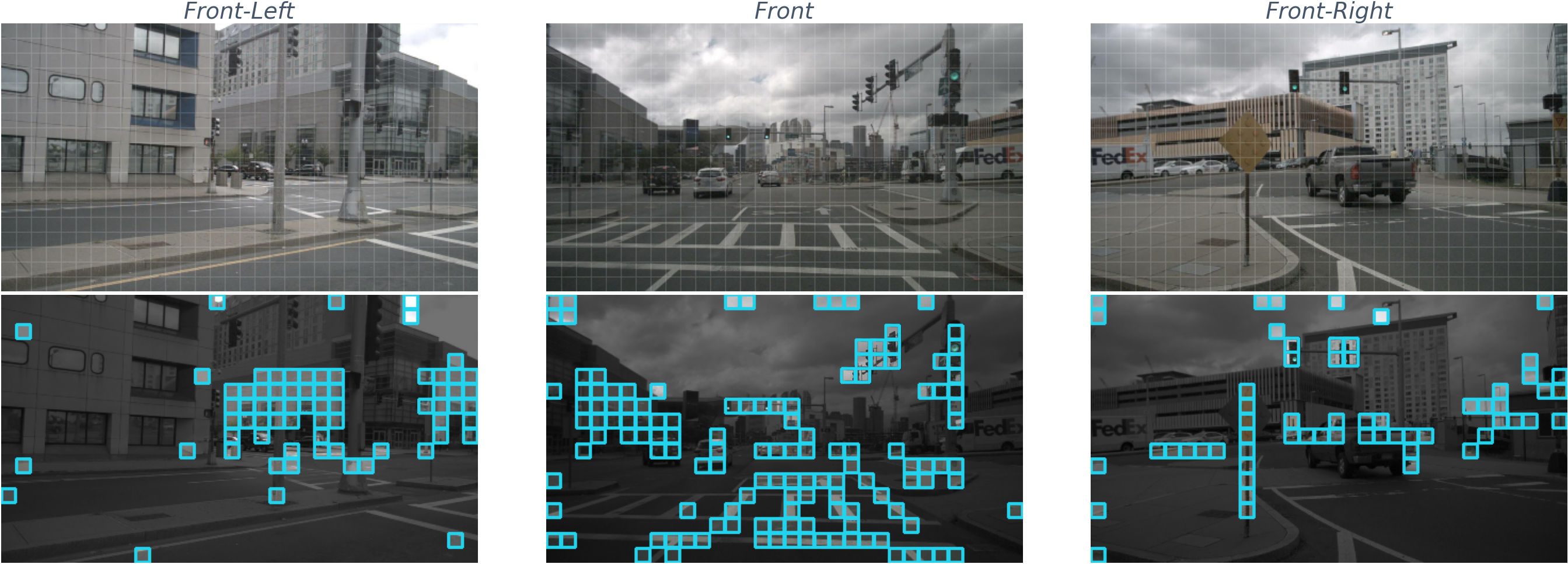}
  \caption{\textbf{\gats at 10\% budget:} \textbf{urban intersection} with FedEx truck, multiple vehicles, and traffic signals.
  Most driving-relevant structure is covered while sky and distant buildings remain pruned.}
  \label{fig:appx_04}
\end{figure*}

\begin{figure*}[t]
  \centering
  \includegraphics[width=0.95\linewidth]{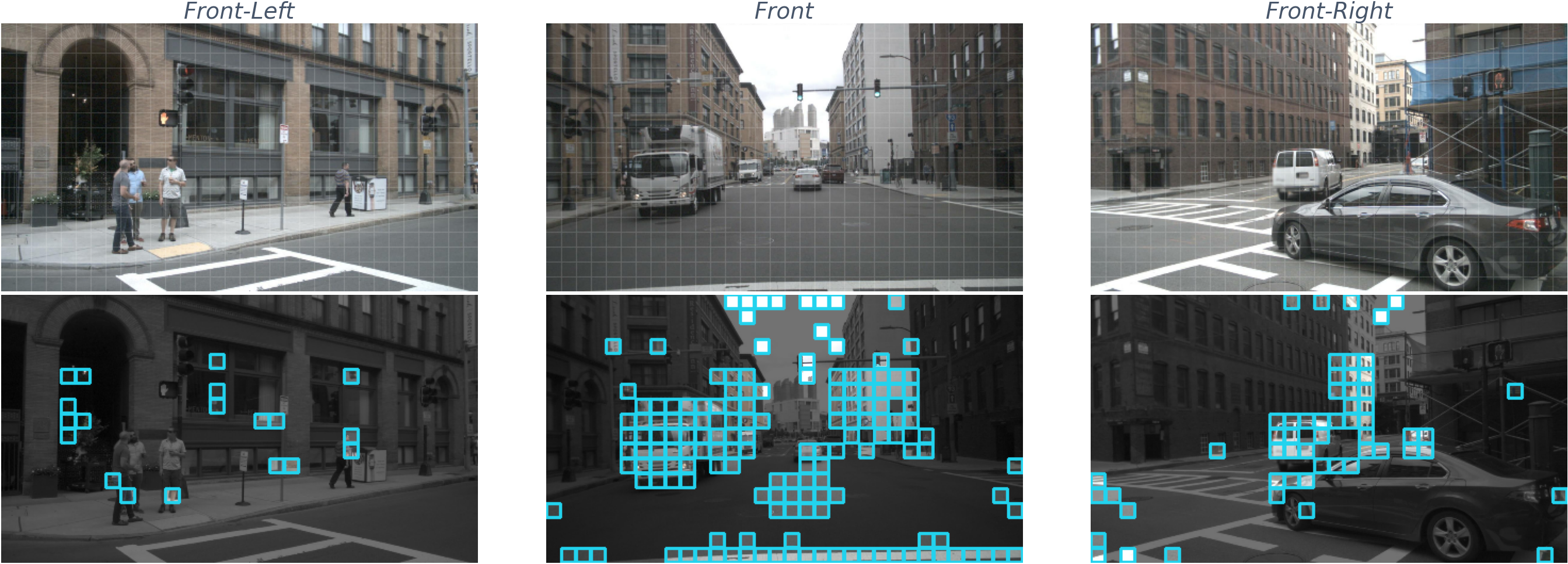}
  \caption{\textbf{\gats at 10\% budget:} \textbf{dense urban traffic} with pedestrians on crosswalk, vehicles, and traffic lights.
  \gats concentrates tokens on all road users while discarding irrelevant background.}
  \label{fig:appx_08}
\end{figure*}

\begin{figure*}[t]
  \centering
  \includegraphics[width=0.95\linewidth]{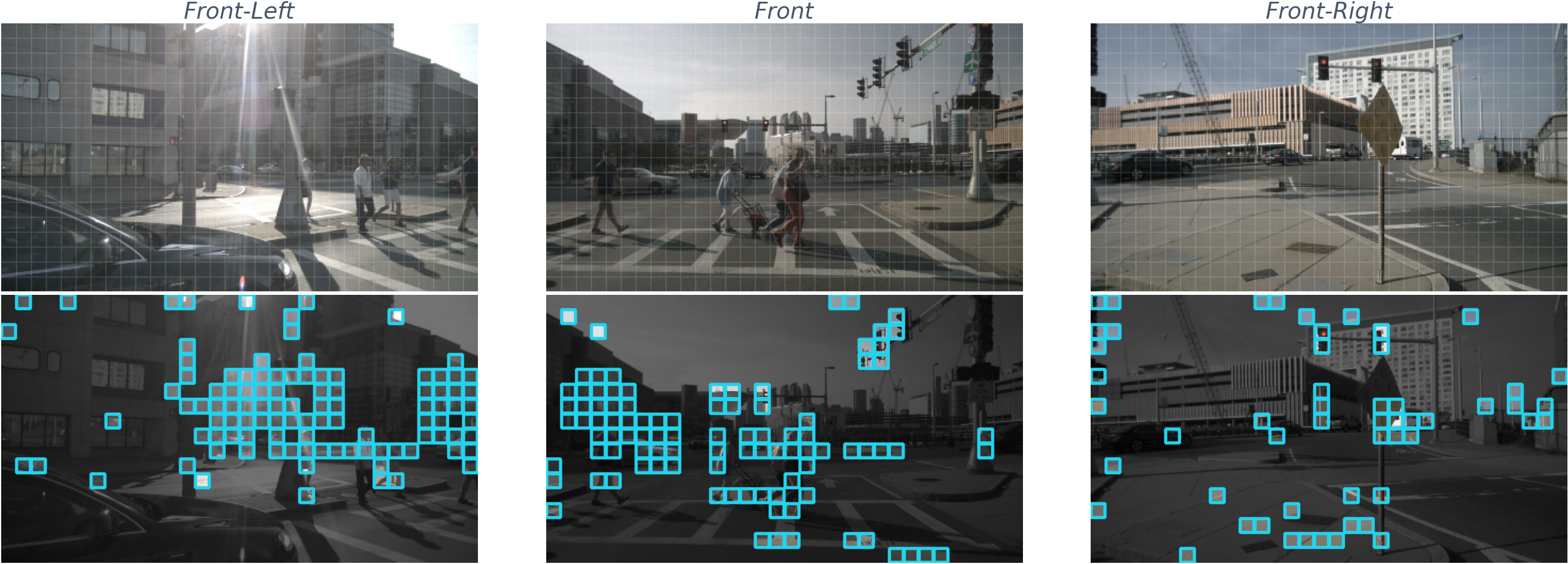}
  \caption{\textbf{\gats at 10\% budget:} \textbf{braking scenario} with multiple pedestrians crossing under backlit conditions.
  \gats selects tokens on pedestrians and nearby vehicles despite challenging lighting.}
  \label{fig:appx_27}
\end{figure*}

\begin{figure*}[t]
  \centering
  \includegraphics[width=0.9\linewidth]{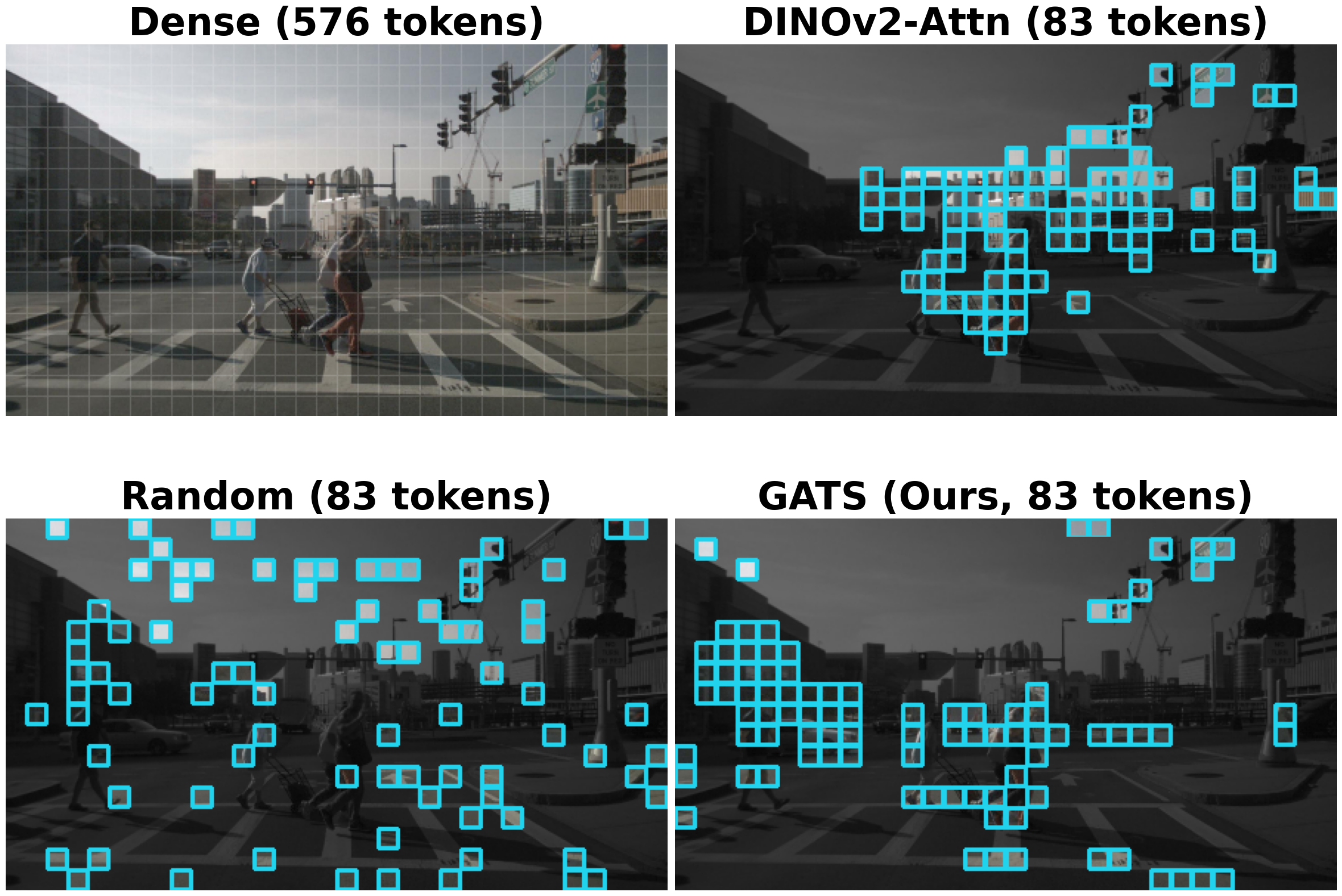}
  \caption{\textbf{Front camera: Dense vs.\ Random vs.\ \gats(10\% budget)} on the braking scenario from \cref{fig:appx_27}.
  Random scatters tokens uniformly; \gats concentrates on pedestrians and the lead vehicle.}
  \label{fig:appx_comp_27}
\end{figure*}

\begin{figure*}[t]
  \centering
  \includegraphics[width=0.9\linewidth]{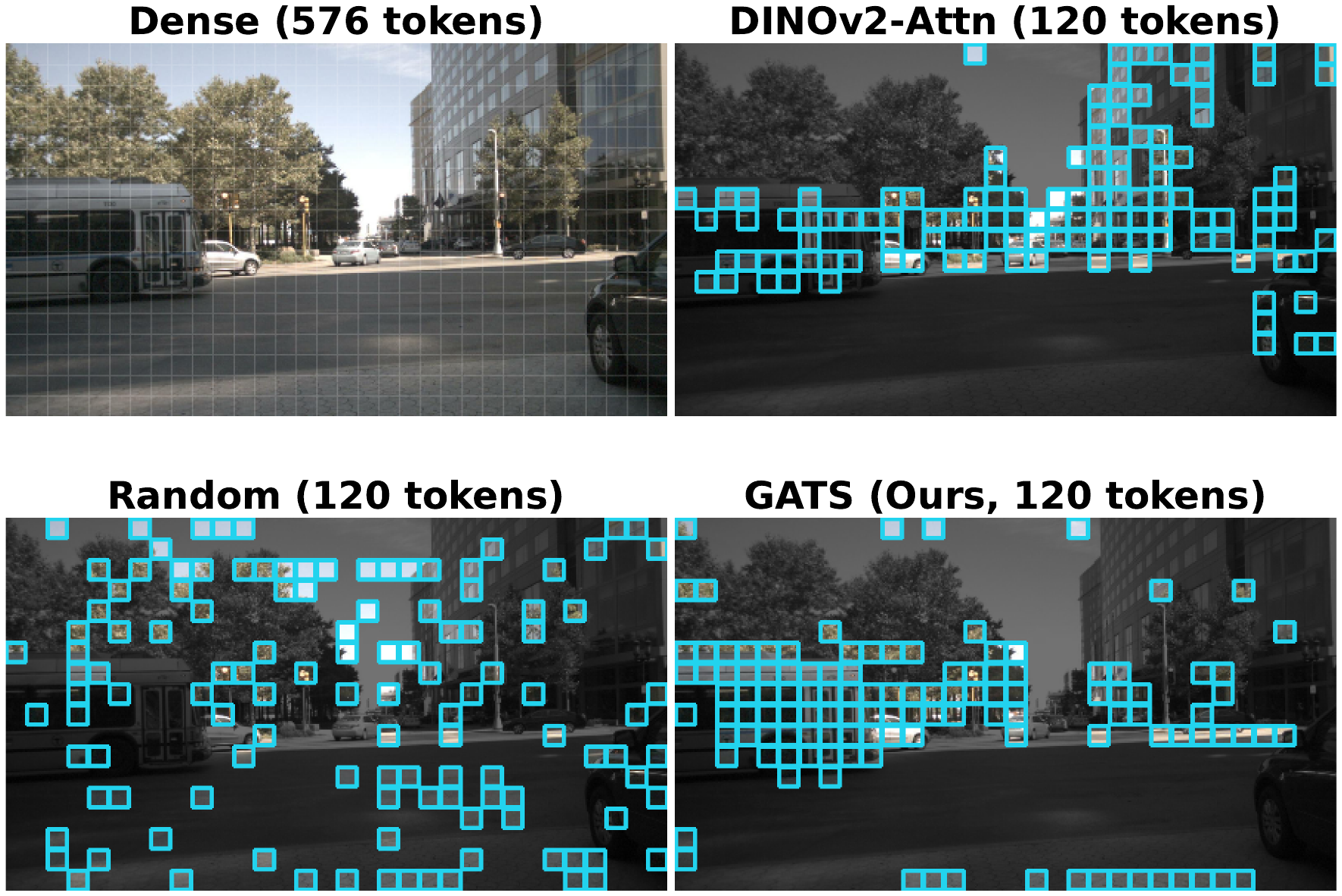}
  \caption{\textbf{Front camera: Dense vs.\ Random vs.\ \gats(10\% budget)} on a scene with a bus and multiple large vehicles.
  \gats covers all large vehicles; Random misses portions of them.}
  \label{fig:appx_comp_30}
\end{figure*}

\begin{figure*}[t]
  \centering
  \includegraphics[width=0.95\linewidth]{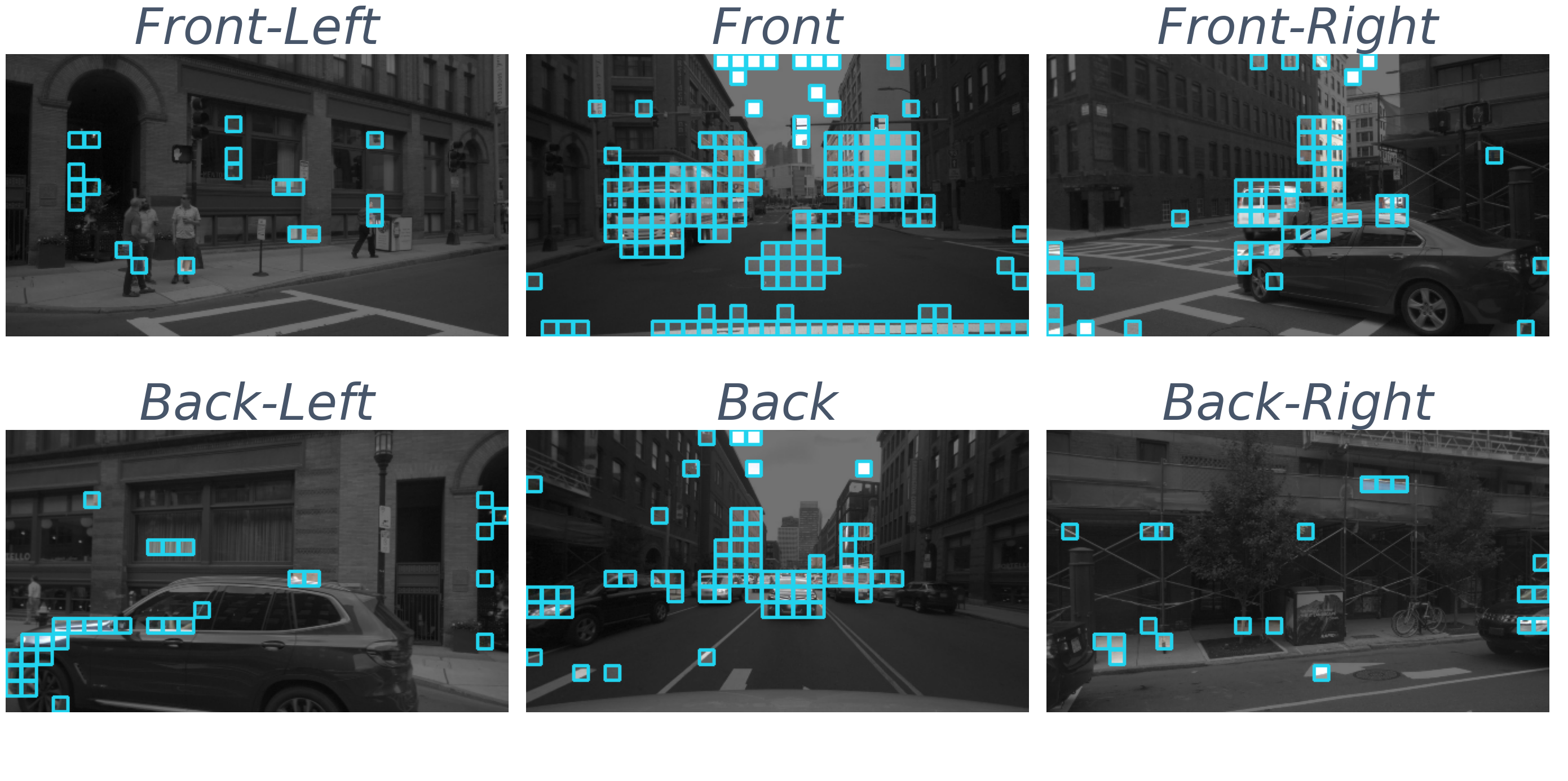}
  \caption{\textbf{Full qualitative result} on the urban traffic scene from \cref{fig:appx_08} (6 cameras).
  \gats(10\% budget) concentrates tokens on all road users while pruning sky and distant buildings.}
  \label{fig:appx_qual_08}
\end{figure*}

\Cref{fig:appx_04,fig:appx_08,fig:appx_27} show token selection at the 10\% budget across diverse urban scenes. \Cref{fig:appx_comp_27,fig:appx_comp_30} compare selection strategies side by side, and \cref{fig:appx_qual_08} provides a full six-camera view.

\subsection{Backbone Analysis}
\label{app:backbone}

\begin{table}[t]
\centering
\small
\caption{\textbf{Backbone comparison} at 10\% budget.
\gats consistently outperforms random and other selectors with different backbones.}
\label{tab:backbone}
\begin{tabular*}{\columnwidth}{@{\extracolsep{\fill}}llcc}
\toprule
Backbone & Selector & L2$_\text{avg}$$\downarrow$ & Coverage$\uparrow$ \\
\midrule
DINOv2-S/14 & \gats & \textbf{0.537} & \textbf{86.3\%} \\
DINOv2-S/14 & DINOv2-Attn & 0.628 & 84.7\% \\
DINOv2-S/14 & Grid & 0.563 & 64.4\% \\
DINOv2-S/14 & Random & 0.560 & 60.2\% \\
\midrule
DINOv3-S & \gats & \textbf{0.550} & \textbf{85.1\%} \\
DINOv3-S & Random & 0.568 & 60.2\% \\
\midrule
ViT-S/16 (sup.) & \gats & \textbf{0.543} & \textbf{83.6\%} \\
ViT-S/16 (sup.) & Random & 0.564 & 60.2\% \\
\bottomrule
\end{tabular*}
\end{table}

\gats consistently outperforms Random under all three backbones in both planning quality and hazard coverage (\cref{tab:backbone}).
The coverage advantage is largest with self-supervised backbones (DINOv2, DINOv3), whose spatially uniform feature distribution rewards informed selection, but remains substantial even with a supervised ViT (coverage: 83.6\% vs.\ 60.2\%).

\subsection{NAVSIM Ablation}
\label{app:navsim_ablation}

NAVSIM ablation results at 10\% budget are shown in \cref{tab:ablation_navsim_10}.
All components contribute consistently: removing the road stream causes the largest drop ($-$8.2\%), followed by the world model ($-$5.8\%) and the static prior ($-$5.3\%).
The 25\% budget NAVSIM ablation shows the same trend and is omitted for space. Notably, unlike nuScenes, random and uniform selectors fall below Dense fusion on NAVSIM (\cref{tab:navsim}). This confirms that geometry-guided routing—not mere token dropping—is essential in complex scenarios.

\subsection{10\% Budget Tables}
\label{app:tables_10}

For reference, we provide the full set of results at the extreme-efficiency 10\% budget point: intent-critical coverage (\cref{tab:intent_10}), motion prediction (\cref{tab:motion_10}), ablation on nuScenes (\cref{tab:ablation_10}) and NAVSIM (\cref{tab:ablation_navsim_10}), and robustness (\cref{tab:robust_10}).

\begin{table}[t]
\centering
\small
\caption{\textbf{Intent-critical analysis} at 10\% budget on 739 mined samples.}
\label{tab:intent_10}
\vspace{-2mm}
\begin{tabular*}{\columnwidth}{@{\extracolsep{\fill}}lcc}
\toprule
Selector ($\beta{=}10\%$) & Cov.$\uparrow$ & Col.$\downarrow$ \\
\midrule
\textbf{GATS} & \textbf{86.3\%} & \textbf{0.09\%} \\
DINOv2-Attn & 84.7\% & 0.18\% \\
Random & 60.2\% & 0.38\% \\
Grid & 64.4\% & 0.26\% \\
Learned~\cite{jaegle2021perceiverio} & 16.5\% & 0.52\% \\
Distance~\cite{lu2024latp} & 10.7\% & 0.64\% \\
\midrule
Dense fusion & 96.8\% & 0.34\% \\
\bottomrule
\end{tabular*}
\vspace{-3mm}
\end{table}

\begin{table}[t]
\centering
\small
\caption{\textbf{Motion prediction} at 10\% budget (EPE in m at 2\,s).}
\label{tab:motion_10}
\vspace{-2mm}
\resizebox{\columnwidth}{!}{%
\begin{tabular}{lcccc}
\toprule
Method & EPE$_\text{mov}$@2s$\downarrow$ & Zero@2s$\uparrow$ & Rel.\%$\uparrow$ & EPE$_\text{sta}$@2s$\downarrow$ \\
\midrule
\textbf{GATS (10\%)} & \textbf{5.10} & \textbf{13.29} & \textbf{61.6\%} & \textbf{0.48} \\
Random (10\%) & 7.22 & 11.93 & 39.5\% & 0.60 \\
Grid (10\%) & 7.16 & 11.88 & 39.8\% & 0.61 \\
Dense fusion & 6.62 & 11.92 & 44.5\% & 0.59 \\
DINOv2-Attn (10\%) & 5.46 & 11.40 & 52.1\% & 0.50 \\
Distance (10\%) & 6.18 & 12.22 & 49.4\% & 0.59 \\
Learned (10\%) & 11.22 & 11.12 & $-$0.9\% & 0.70 \\
Pool (10\%) & 8.82 & 12.25 & 28.0\% & 0.74 \\
\bottomrule
\end{tabular}%
}
\vspace{-3mm}
\end{table}

\begin{table}[t]
\centering
\small
\caption{\textbf{Ablation} at 10\% budget on nuScenes val.}
\label{tab:ablation_10}
\vspace{-2mm}
\begin{tabular*}{\columnwidth}{@{\extracolsep{\fill}}lccc}
\toprule
Variant & L2$_\text{avg}$$\downarrow$ & $\Delta_\text{L2}$ & Col\%$\downarrow$ \\
\midrule
Full (GATS 10\%) & \textbf{0.537} & -- & \textbf{0.09} \\
\midrule
w/o radar velocity & 0.602 & +12.1\% & 0.15 \\
w/o motion residuals & 0.558 & +3.9\% & 0.14 \\
w/o 3D anchoring & 0.601 & +11.9\% & 0.10 \\
w/o past flow & 0.592 & +10.2\% & 0.10 \\
w/o learned refine. & 0.565 & +5.1\% & 0.10 \\
w/o road stream & 0.584 & +8.7\% & 0.11 \\
w/o static prior & 0.556 & +3.5\% & 0.10 \\
w/o world model & 0.625 & +16.4\% & 0.28 \\
\bottomrule
\end{tabular*}
\vspace{-3mm}
\end{table}

\begin{table}[t]
\centering
\small
\caption{\textbf{Ablation} at 10\% budget on NAVSIM navtest.}
\label{tab:ablation_navsim_10}
\vspace{-2mm}
\begin{tabular*}{\columnwidth}{@{\extracolsep{\fill}}lccc}
\toprule
Variant & PDMS$\uparrow$ & EPDMS$\uparrow$ & $\Delta_\text{EPDMS}$ \\
\midrule
Full (GATS 10\%) & \textbf{88.9} & \textbf{87.9} & -- \\
\midrule
w/o motion residuals & 86.0 & 85.0 & $-$3.3\% \\
w/o 3D anchoring & 84.6 & 83.4 & $-$5.1\% \\
w/o past flow & 84.9 & 83.9 & $-$4.6\% \\
w/o learned refine. & 84.5 & 83.5 & $-$5.0\% \\
w/o road stream & 81.7 & 80.7 & $-$8.2\% \\
w/o static prior & 84.5 & 83.2 & $-$5.3\% \\
w/o world model & 84.0 & 82.8 & $-$5.8\% \\
\bottomrule
\end{tabular*}
\vspace{-3mm}
\end{table}

\begin{table}[t]
\centering
\small
\caption{\textbf{Robustness} at 10\% budget (severity 3; same 5 corruption types as \cref{tab:robust}).}
\label{tab:robust_10}
\vspace{-2mm}
\resizebox{\columnwidth}{!}{%
\begin{tabular}{lcccc}
\toprule
Method & Clean L2$\downarrow$ & Corrupt L2$\downarrow$ & Rel.\ degrad.$\downarrow$ & Mot.\ blur$\downarrow$ \\
\midrule
Dense camera & 0.668 & 0.924 & +38.3\% & +130\% \\
Dense fusion & 0.573 & 0.723 & +26.2\% & +88\% \\
\midrule
\textbf{GATS (10\%)} & \textbf{0.537} & \textbf{0.610} & +13.6\% & +31\% \\
Random (10\%) & 0.560 & 0.624 & +11.4\% & +27\% \\
Grid (10\%) & 0.563 & 0.637 & +13.2\% & +29\% \\
DINOv2-Attn (10\%) & 0.628 & 0.717 & +14.2\% & +31\% \\
Learned (10\%) & 0.669 & 0.763 & +14.1\% & +34\% \\
Distance (10\%) & 0.675 & 0.740 & +9.7\% & +28\% \\
\bottomrule
\end{tabular}%
}
\vspace{-3mm}
\end{table}

\subsection{Implementation Details}
\label{app:impl}

\paragraph{LiDAR preprocessing.}
We use up to 5 LiDAR sweeps (0.25\,s history).
Ground segmentation uses RANSAC on the lowest 10\% of points per BEV grid cell.
Motion residuals are computed by accumulating point-wise displacements after compensating for ego motion.
Radar detections are projected using their 3D positions and radial velocities.

\paragraph{Token selection.}
For the obstacle stream, we use a 2-layer MLP (hidden dim 32, ReLU) on the 16-dimensional geometric feature vector $\mathbf{g}_i$.
The road stream uses a similar MLP on ground-point features (intensity, height variance, point count).
The visual reserve uses a 1-layer MLP on the DINOv2 [CLS] attention weights during dense-backbone training, and falls back to a geometry-based heuristic during sparse-backbone deployment.

\paragraph{World model training.}
The JEPA-style feature prediction loss weight is 1.0; match loss is 0.5; anchor loss is 0.1; static regularization is 0.01.
Match radius $r$ in \cref{eq:match} is 3.0\,m.
We use 8 random query patches per step for the feature prediction head.

\paragraph{Planning head.}
The regression head uses 2 cross-attention layers ($d{=}256$, 8 heads) followed by a 3-layer MLP.
The vocabulary head projects the same query to a $V{=}256$ logit vector.
Trajectory prototypes are obtained by $k$-means on training trajectories.

\paragraph{Sparse backbone distillation.}
Phase 1 (3 epochs): only the backbone is trained; loss = MSE between sparse and dense DINOv2 features at selected patches.
Phase 2 (3 epochs): joint training with backbone MSE loss ($\lambda{=}0.5$) and the full planning + world-model loss.
Learning rates are $10^{-4}$ and $3{\times}10^{-5}$ respectively, with cosine decay.

\paragraph{Code availability.}
Code and pretrained checkpoints will be made publicly available upon acceptance.

\end{document}